\documentclass[conference]{IEEEtran}
\usepackage{times}

\usepackage[numbers]{natbib}
\usepackage{multicol}
\usepackage[bookmarks=true]{hyperref}

\usepackage{epsfig}
\usepackage{graphicx}
\usepackage{amsmath}
\usepackage{amssymb}

\usepackage{subfigure}

\usepackage{interval}

\usepackage{caption}

\usepackage{makecell}
\usepackage{booktabs}
\usepackage{multirow}
\usepackage{siunitx}

\usepackage[T1]{fontenc}
\usepackage{fix-cm}

\usepackage{times}
\usepackage{epsfig}
\usepackage{graphicx}
\usepackage{amsmath}
\usepackage{amssymb}

\usepackage{array}

\usepackage{caption}

\usepackage{makecell}
\usepackage{booktabs}
\usepackage{multirow}
\usepackage{siunitx}

\usepackage{fix-cm}

\newcolumntype{?}{!{\vrule width 1pt}}
\newcolumntype{C}[1]{>{\centering}m{#1}}

\newcolumntype{X}{@{\hskip\tabcolsep\vrule width 1pt\hskip\tabcolsep}}

\newcommand{\myfigurefoursmallcol}[1]{
\begin{minipage}[b]{.1\textwidth}
\includegraphics[width=1.125\linewidth]{#1}
\end{minipage}
}

\newcommand{\myfigurefoursmallcolcaption}[2]{
\begin{minipage}[b]{.1\textwidth}
\includegraphics[width=1.125\linewidth]{#1}
\caption{{\scriptsize {#2}}}
\end{minipage}
}

\newcommand{\myfiguresixcol}[1]{
\begin{minipage}[b]{.14\textwidth}
\includegraphics[width=1.075\linewidth]{#1}
\end{minipage}
}

\newcommand{\myfiguresixcolcaption}[2]{
\begin{minipage}[b]{.14\textwidth}
\includegraphics[width=1.075\linewidth]{#1}
\caption{{\small {#2}}}
\end{minipage}
}

\begin{document}

\title{First-Person Action-Object Detection with EgoNet}

\author{Gedas Bertasius$^{1}$, Hyun Soo Park$^2$, Stella X. Yu$^3$, Jianbo Shi$^1$\\
$^1$University of Pennsylvania, $^2$University of Minnesota, $^3$UC Berkeley ICSI\\
{\tt\small \{gberta,jshi\}@seas.upenn.edu} \ \ \ {\tt\small hspark@umn.edu}  \ \ \ {\tt\small stella.yu@berkeley.edu}
}




%

\maketitle

\begin{abstract}
Unlike traditional third-person cameras mounted on robots, a first-person camera, captures a person's visual sensorimotor object interactions from up close. In this paper, we study the tight interplay between our momentary visual attention and motor action with objects from a first-person camera. We propose a concept of action-objects---the objects that capture person's conscious visual (watching a TV) or tactile (taking a cup) interactions. Action-objects may be task-dependent but since many tasks share common person-object spatial configurations, action-objects exhibit a characteristic 3D spatial distance and orientation with respect to the person. 

We design a predictive model that detects action-objects using EgoNet, a joint two-stream network that holistically integrates visual appearance (RGB) and 3D spatial layout (depth and height) cues to predict per-pixel likelihood of action-objects. Our network also incorporates a first-person coordinate embedding, which is designed to learn a spatial distribution of the action-objects in the first-person data. We demonstrate EgoNet's predictive power, by showing that it consistently outperforms previous baseline approaches. Furthermore, EgoNet also exhibits a strong generalization ability, i.e., it predicts semantically meaningful objects in novel first-person datasets. Our method's ability to effectively detect action-objects could be used to improve robots' understanding of human-object interactions.

\end{abstract}


\IEEEpeerreviewmaketitle

\section{Introduction}

\begin{figure}[t]
\begin{center}
   \includegraphics[width=1.025\linewidth]{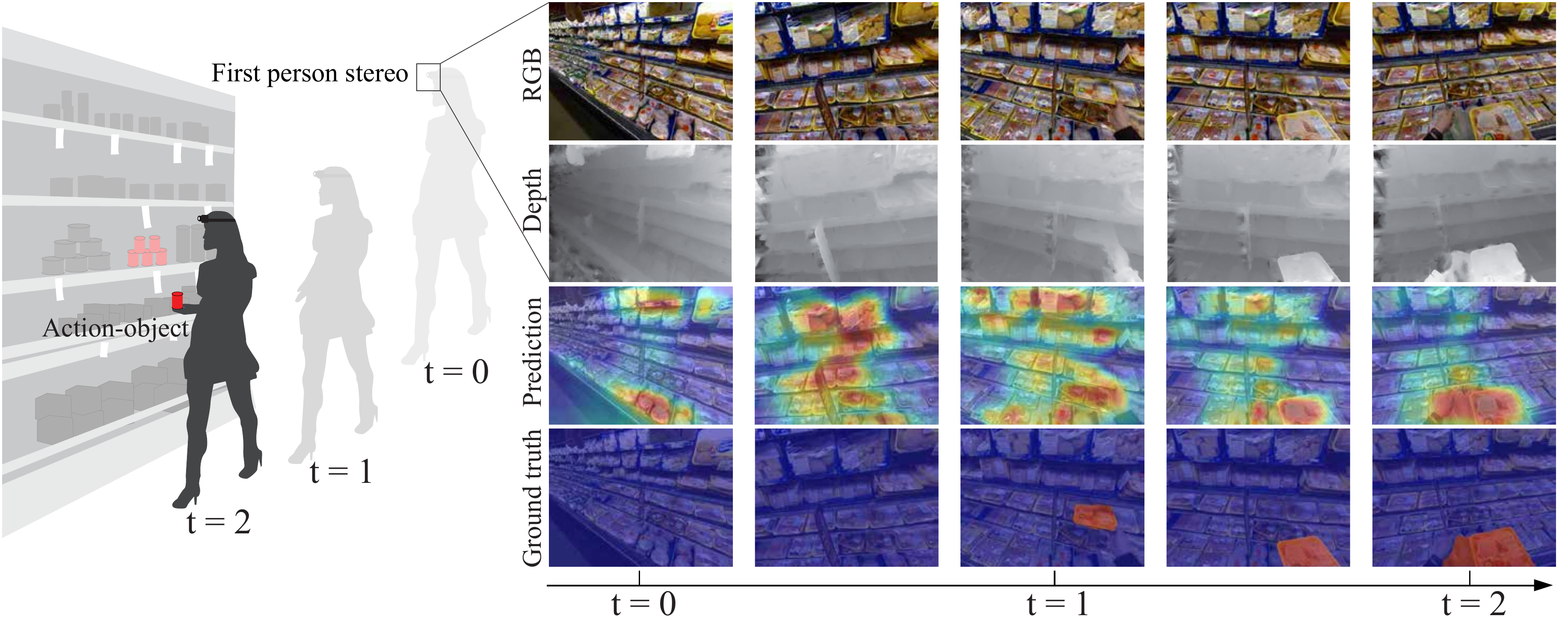}
\end{center}
\vspace{-0.4cm}
   \caption{We predict action-objects from first-person RGBD images (best viewed in color) where action-objects are defined as objects that facilitate people's conscious tactile (grabbing a food package) or visual interactions (watching a TV). Left: a woman approaches a shelf to pick up a food item (red). Right: The food (action-object) is detected progressively as she approaches and reaches her hand to pick it up.\vspace{-0.5cm}}
\label{main_fig}
\end{figure}

Our visual sensation is developed along with the neuromotor system while interacting with surrounding objects~\cite{johansson:2001,perone:2008,vidoni:2009,bowman:2009,lazzari:2009}. As the visual sensation and motor signal reinforce each other, it characterizes the way we progressively interact with objects in 3D, which provides a strong cue for robotic agents to identify the objects in action among many surroundings. For instance, consider a woman entering a canned food corner at a grocery store as shown in Figure~\ref{main_fig}. When she schemes through hundreds of canned foods to find the tuna can that she looks for, she remains 3-5m from the food stand for efficient search. Once she finds the tuna, she approaches it (1-3m), and then reaches her left hand to pick the tuna can ($<$1m). While she gazes at the expiration date in the label of the can, the distance gets smaller ($<$0.5m). Not only does the tuna can stimulate her visual attention but it also affects her physical actions, such as head or hand movements. Can an infrastructured robots in the grocery store such as Amazon Go\footnote{\url{https://www.amazon.com/b?node=16008589011}} identify the objects in action from the sequence of her actions?

We define such object as an action-object---an object that triggers conscious visual and motor signals. The key properties of an action-object are: (1) it facilitates a person's tactile (touching a cup) or (2) visual (watching a TV) interactions and (3) it exhibits a characteristic distance and orientation with respect to the person. These properties provide strong cues to predicting person's behavior, which allow robots to timely respond to it. 

While in-situ wearable sensors such as gaze tracker with EEG measurements or tactile and force/torque sensors for muscle movement are viable solutions to identify action-objects more accurately than distant sensors such as third-person robot mounted cameras, their integration into our daily life is highly limited. A fundamental question is ``can we detect action-objects as we observe the person interacting with her/his environment from a first-person video alone?''.  This is challenging despite recent success of robot/computer vision systems because (a) a person's gaze direction does not necessarily correspond to action-objects. In other words, not all objects within the person's field of view are consciously attended; (b) action-objects are often task-dependent, which makes it difficult to detect them without knowing the task beforehand; (c) action-objects are not specific to object category, i.e., many object categories correspond to the same action, e.g., TV and a mirror both afford a seeing action. Therefore, an object specific model cannot represent action-objects.

\begin{table*}[ht]
\scriptsize
\centering
 \begin{tabular}{ c X  p{1.4cm} | p{1.4cm} | p{1.4cm} | p{1.4cm} | p{1.4cm} | p{1.4cm} | p{1.4cm} |}
\hline

Scene & cooking & hotel & grocery & desk work & dining & shopping & dishwashing \\
\hline
Frames & 1030 & 410 & 463 & 491 & 515 & 646 & 674\\
\hline
 RGB & \includegraphics[width=1.38cm]{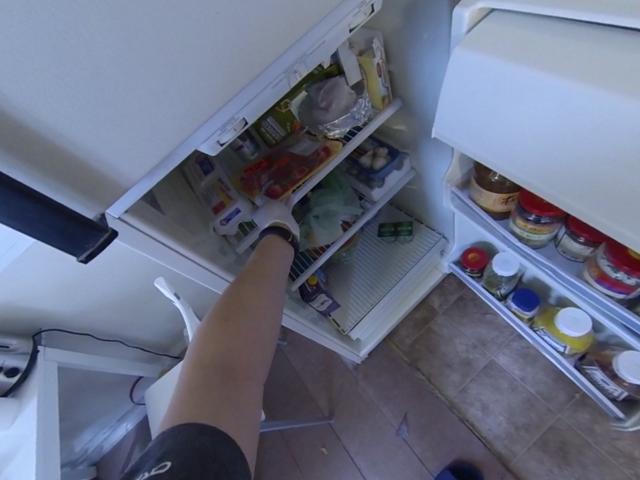} & \includegraphics[width=1.38cm]{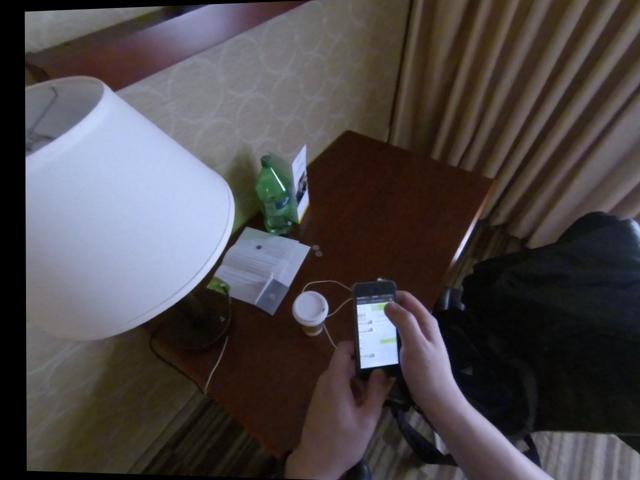} & \includegraphics[width=1.38cm]{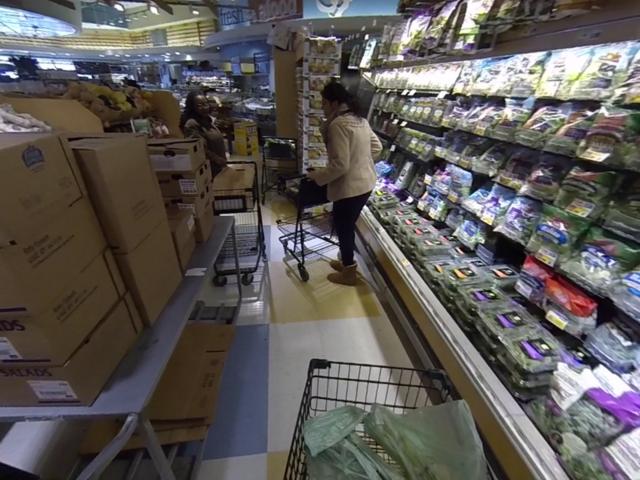} & \includegraphics[width=1.38cm]{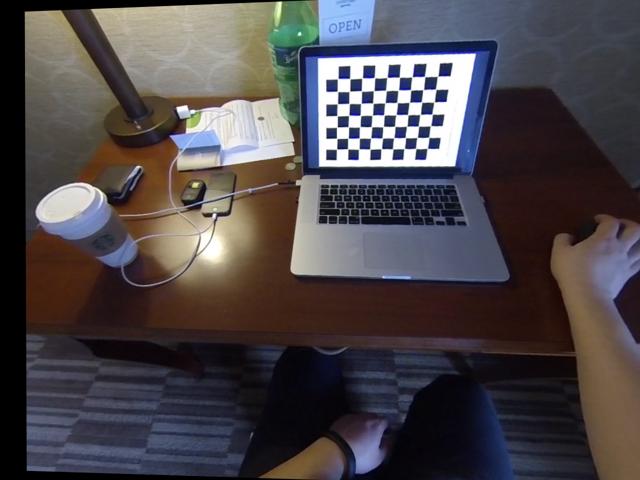} & \includegraphics[width=1.38cm]{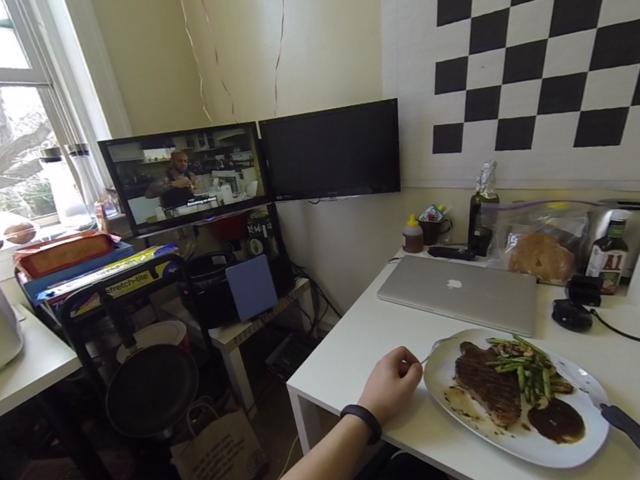} & \includegraphics[width=1.38cm]{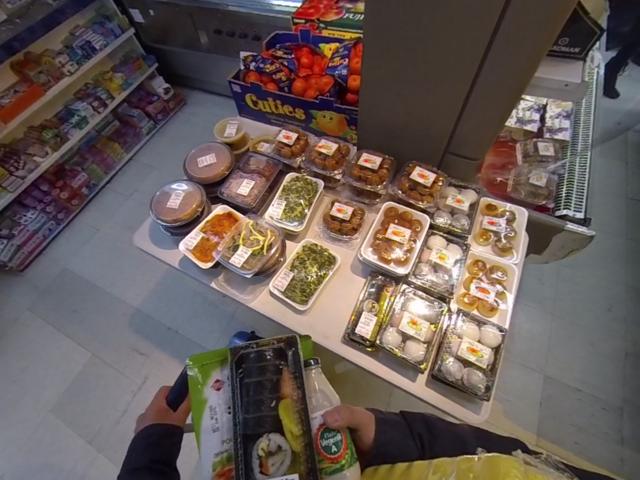} & \includegraphics[width=1.38cm]{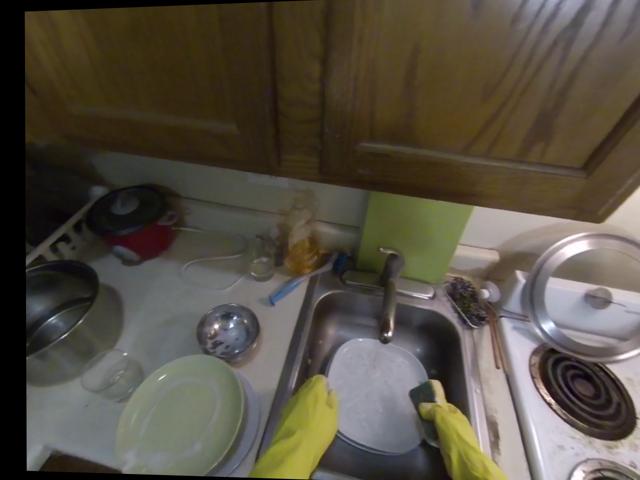}\\ \hline
DHG & \includegraphics[width=1.38cm]{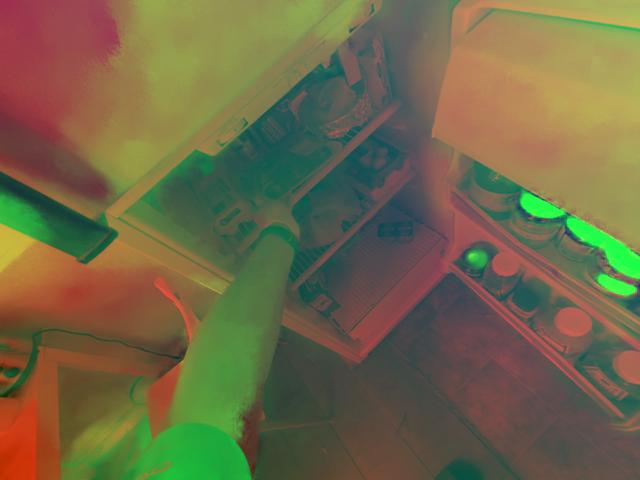} & \includegraphics[width=1.38cm]{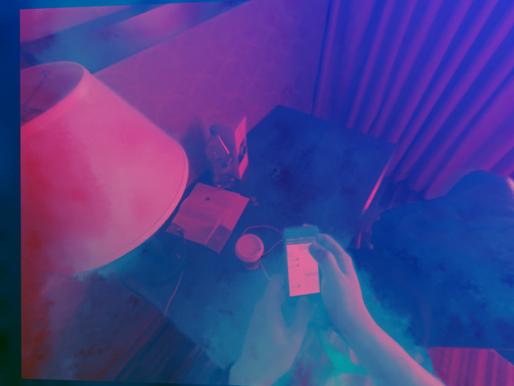} & \includegraphics[width=1.38cm]{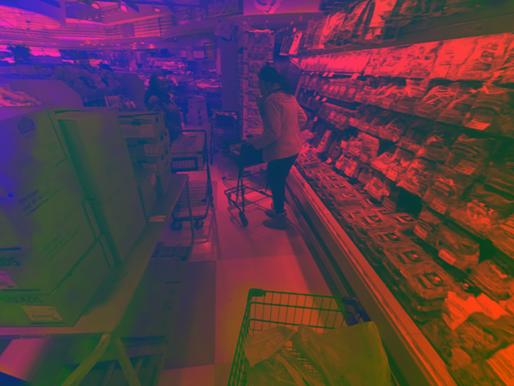} & \includegraphics[width=1.38cm]{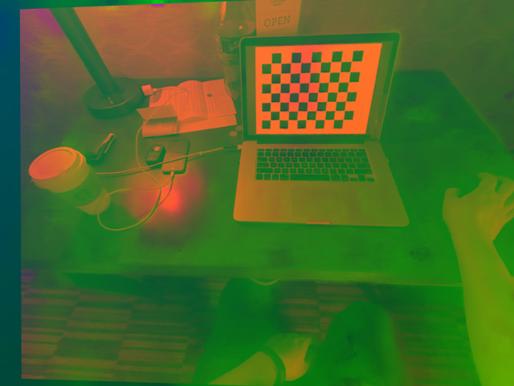} & \includegraphics[width=1.38cm]{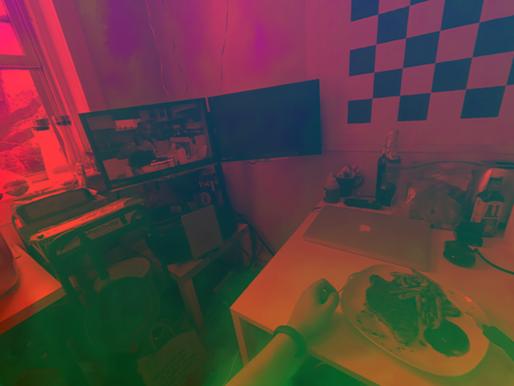} & \includegraphics[width=1.38cm]{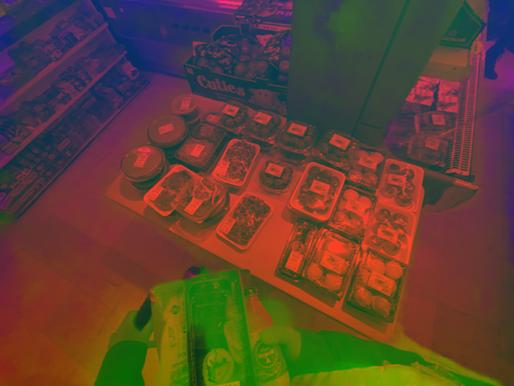} & \includegraphics[width=1.38cm]{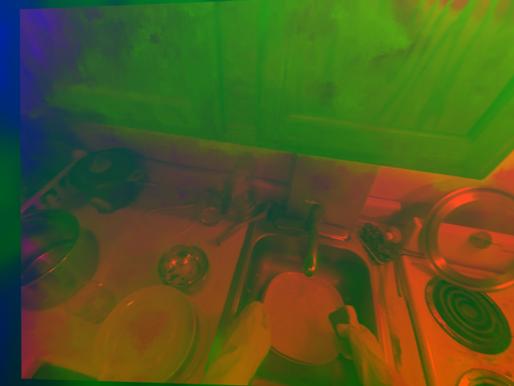}\\ \hline
Ground Truth & \includegraphics[width=1.38cm]{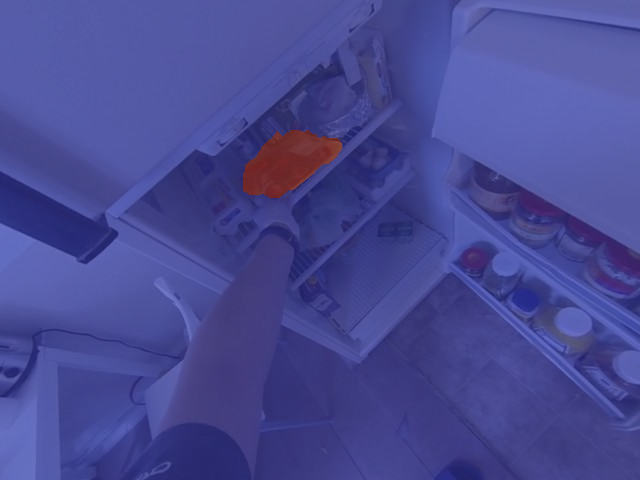} & \includegraphics[width=1.38cm]{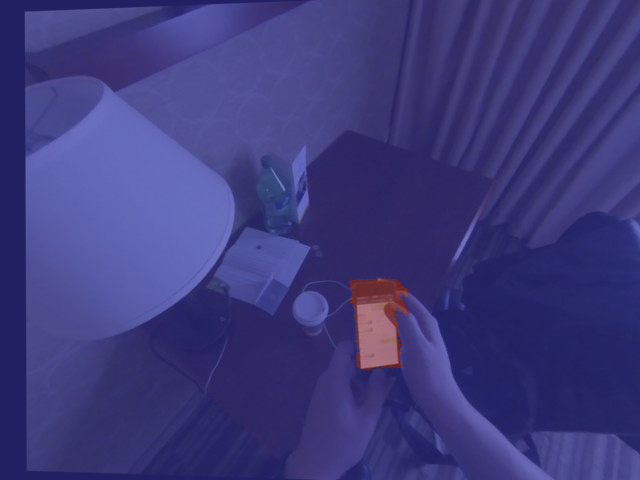} & \includegraphics[width=1.38cm]{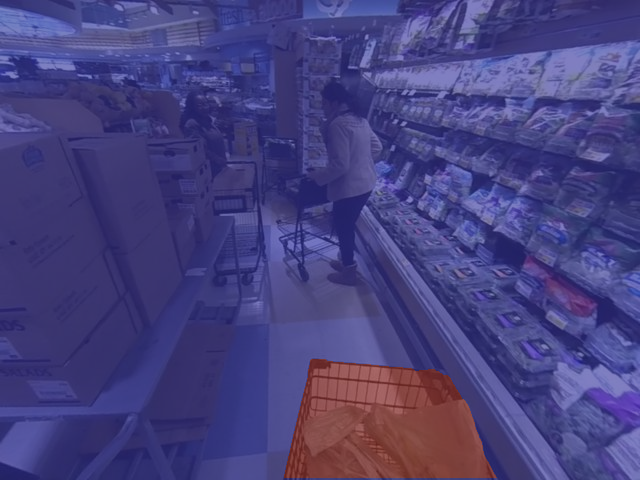} & \includegraphics[width=1.38cm]{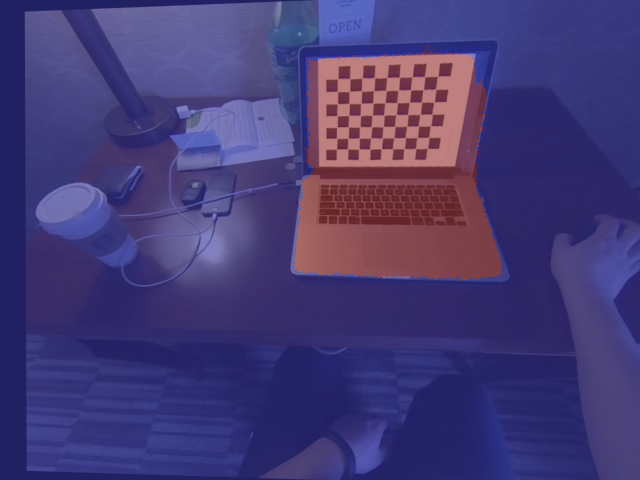} & \includegraphics[width=1.38cm]{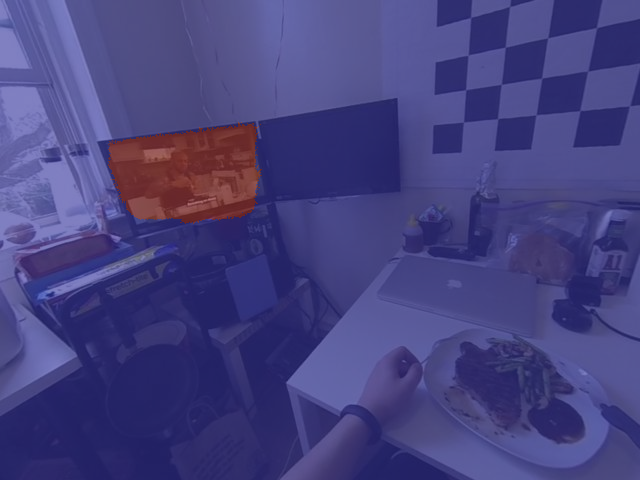} & \includegraphics[width=1.38cm]{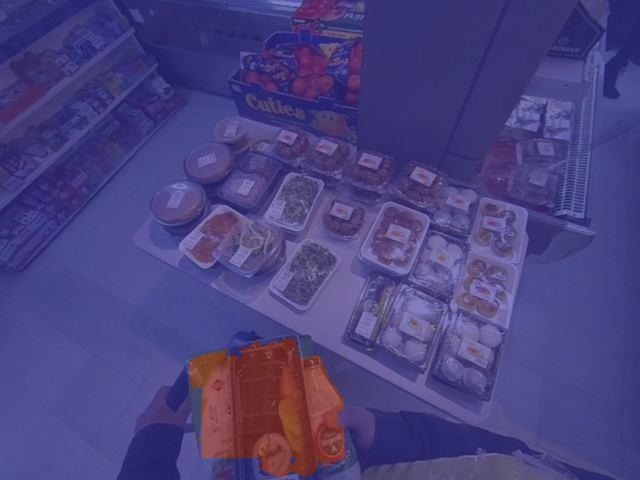} & \includegraphics[width=1.38cm]{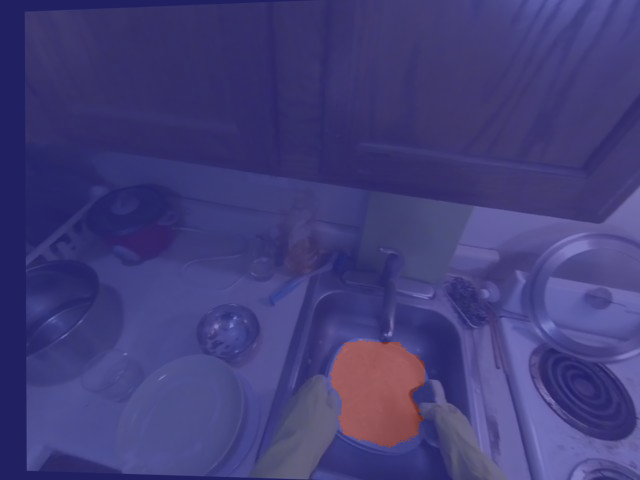}\\ 
\hline

\end{tabular}
\caption[First Person Action-Object Dataset]{The summary of our First Person Action-Object RGBD dataset, which captures people performing $7$ different activities. Our dataset contains $4247$ frames with RGB, and DHG (depth, height, grayscale image) inputs as well as annotated per-pixel action-object masks.\vspace{-0.5cm}}\label{table:dataset}
\end{table*}

In this paper, we address these challenges by leveraging a first-person stereo camera system and our proposed EgoNet model, a joint two-stream network that integrates visual appearance (RGB) and 3D spatial cues (depth and height). These two pathways are complementary: one of which learns visual appearance cues, while the other exploits 3D spatial information indicative of action-objects. These are combined via a joint pathway, which incorporates a first-person coordinate embedding that learns an action-object spatial distribution in the first-person image. The entire network is jointly optimized to the action-object ground truth that is provided by the camera wearer. We quantitatively justify the architecture choices of our EgoNet model and show that it outperforms all prior approaches for the action-object detection task across multiple first-person datasets. We also show that EgoNet generalizes well on a variety of novel datasets, even without being adapted to a specific task as is commonly done~\cite{Li_2015_CVPR,conf/cvpr/RenG10,ma2016going}.


In conjunction with the EgoNet, we present a new first-person action-object RGBD dataset that includes object interactions during diverse activities such as cooking, shopping, dish-washing, working, etc. The camera wearer who is aware of the task and who can disambiguate conscious visual attention and subconscious gaze activities provides per-pixel binary labels of first-person images, which we then use to build our action-object model in a form of EgoNet.

Our framework is different from a classic object detection task because action-objects are associated with actions without explicit object categories. It also differs from a visual saliency detection because visual saliency does not necessarily correspond to a specific action. Finally, our action-object task differs from activity recognition because we detect action-objects by exploiting common person-object spatial configurations instead of modeling activity-specific interactions, which makes our model applicable to different activities.

\textbf{Why Robotics and First-Person?} Precisely identifying action-objects is a fundamental task in human-robot interaction where a robotic system measures the internal state of humans and tries to answer questions such as ``what is that person doing?'', ``what will he do next?'', ``how can I assist him?''. However, answering these questions from a third-person perspective is often challenging because a person's actions are captured from a relatively large distance and possibly from a suboptimal orientation, which makes it difficult to recognize that person's actions and understand his intentions. Also a first-person view contains inherent task-intention via a person's head and body orientation relative to the objects, and therefore, the task information does not need to be modeled explicitly using action recognition as is commonly done in prior work~\cite{ma2016going,Li_2015_CVPR}.

\section{Related Work}


\textbf{Complementary Object and Activity Recognition in Third-Person.} Actions are performed in the context of objects. This coupling provides a complementary cue to recognize actions. Wu et al.~\cite{wu:2007} leveraged object information to classify fine-grained activities. Yao and Fei-Fei~\cite{yao:2010,yao:2010_a} have presented a spatial model between human pose and objects for activity recognition. Some approaches also used low level bag-of-feature models to learn the spatial relationship between objects and activities from a single third-person image~\cite{delaitre:2011}. Conversely, the activity can provide a functional cue to recognize objects~\cite{matthew:2010,juergen:2011,kjellstrom:2011}. Such a model becomes even more powerful when incorporating the cues of how the object is physically manipulated~\cite{gupta:2007,gupta:2011,rabinovich:2007}. In addition, object affordance can be learned by simulating human motion in the 3D space~\cite{fouhey:2012,yu:2015}. Furthermore, Gori et al.~\cite{DBLP:journals/corr/GoriAR15} proposed to recognize activities from a third-person robot's view using people detections. 

Whereas most of these methods require detected people or body pose as an input, our work leverages a first-person view and does not need to detect people or human pose a priori.

  \begin{figure*}[t]
\centering
  \includegraphics[width=0.8\linewidth]{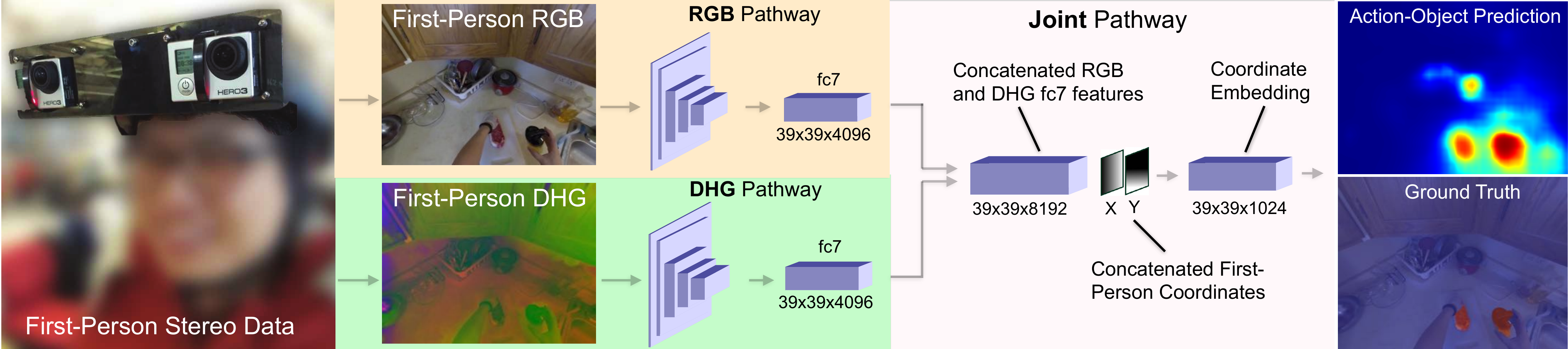}
   \caption{Our proposed EgoNet architecture (best viewed in color) takes as input first-person RGB and DHG images, which encode 2D visual appearance and 3D spatial cues respectively. The fully convolutional RGB pathway then uses the visual appearance cues, while the fully convolutional DHG pathway exploits 3D spatial information to detect action-objects. The information from both pathways is combined via the joint pathway, which also implements the first-person coordinate embedding, and then outputs a per-pixel action-object probability map.\vspace{-0.5cm} }
\label{arch}
\end{figure*}
 
\textbf{First-Person Object Detection.} There exist multiple prior methods that explore object detection from first-person images as a main task~\cite{conf/cvpr/RenG10,conf/cvpr/FathiRR11}, or as an auxiliary task for activity recognition~\cite{PirsiavashR_CVPR_2012_1,Li_2015_CVPR,ma2016going,Fathi:2011:UEA:2355573.2356302,Ryoo:2015:RAP:2696454.2696462} or video summarization~\cite{DBLP:journals/ijcv/LeeG15,Lu:2013:SSE:2514950.2516026}. Below we summarize how our action-object detection task is different from this prior work.

The work in~\cite{Li_2015_CVPR,DBLP:conf/iccv/LiFR13} attempts to predict gaze from the first-person images and use it for activity recognition. However, we know that a person's gaze direction does not always correspond to action-objects but instead capture noisy eye movement patterns, which may not be useful for activity recognition. In the context of our problem, the camera wearer who was performing the task and who can disambiguate conscious visual attention and subconscious gaze activities provides per-pixel binary labels of the first-person images, which we then use to build our action-object model.

  
The methods in~\cite{PirsiavashR_CVPR_2012_1,Li_2015_CVPR} perform object detection and activity recognition disjointly: first an object detector is applied to find all objects in the scene, and then those detections are used for activity recognition without necessarily knowing, which objects the person may be interacting with.  Furthermore, these methods employ a set of predefined object classes. However, many object categories can correspond to the same action, e.g., TV and a mirror both afford a seeing action, and thus, an object class specific model may not be able to represent the action-objects accurately.

Some prior work focused specifically on handled object detection~\cite{conf/cvpr/RenG10,DBLP:conf/rss/CaiKS16}. However, action-object detection task also requires detecting conscious visual interactions that do not necessarily involve hand manipulation (e.g. watching a TV). Furthermore, from a development point of view, conscious visual attention is one way for a person to interact. For instance, for the babies who lack motor skills, their conscious visual attention is the only thing that indicates their action-objects, and thus detecting only handled objects is not enough.

The most relevant to our action-object detection task is the work in~\cite{ma2016going}  that detects objects of interest for activity recognition~\cite{ma2016going}. However, this method is designed specifically for recognizing various cooking activities, which requires detecting mostly handled-objects as in~\cite{conf/cvpr/RenG10}. Thus, the authors~\cite{ma2016going} manually bias their method to recognize objects near hands. Such an approach may not work for other types of activities that do not involve hand manipulation such as watching a TV, interacting with a person, etc.

Finally, none of the above methods, fully exploit common person-object spatial configuration. We hypothesize that a first-person view contains inherent task-intention via a person's head and body orientation relative to the objects. In other words, during an interaction with an object, people position themselves at a certain distance and orientation relative to that object. Thus, 3D spatial information provides essential cues that could be used to recognize action-objects. We leverage such 3D cues, by using first-person stereo cameras, and building an EgoNet model that uses 3D depth and height cues to reason about action-objects.

\textbf{Novelty of Action-Object Concept.} We acknowledge that our defined concept of action-objects overlaps with several concepts from prior work such as object-action complexes (OAC)~\cite{journals/ras/KrugerGPPSWUAKOAD11}, handled-objects~\cite{conf/cvpr/RenG10,DBLP:conf/rss/CaiKS16}, objects-in-action~\cite{gupta:2007}, or object affordances~\cite{kjellstrom:2011}. However, we point out that these prior methods typically focus exclusively on physically manipulated objects, that are specific to certain tasks (e.g. cooking). Instead, the concept of action-objects requires detecting  not only tactile but also conscious visual interactions with the objects (e.g. watching a TV), without making any a-priori assumptions about the task that the person will be performing as is commonly done in prior work~\cite{conf/cvpr/RenG10,DBLP:conf/rss/CaiKS16}.



\textbf{Structured Prediction in First-Person Data.} A task such as action-object detection or visual saliency prediction requires producing a dense probability output for every pixel. To achieve this goal most prior first-person methods employed a set of hand-crafted features combined with a probabilistic or discriminative classifier. For instance, the work in~\cite{DBLP:journals/ijcv/LeeG15} uses manually engineered set of egocentric features with a linear regression classifier to assign probabilities to each region in a segmented image. The method in~\cite{DBLP:journals/corr/BertasiusPS15} exploits the combination of geometric and egocentric cues and trains random forest classifier to predict saliency in first-person images. The work in~\cite{conf/cvpr/RenG10} uses optical flow cues and Graph Cuts~\cite{Boykov:2001:FAE:505471.505473} to compute handled-object segmentations, whereas~\cite{conf/cvpr/FathiRR11} employs transductive SVM to compute foreground segmentation in an unsupervised manner. Finally, some prior work~\cite{DBLP:conf/iccv/LiFR13} integrates a set of hand-crafted features in the graphical model to predict per pixel probabilities of camera wearer's gaze.

We note that the recent introduction of the fully convolutional networks (FCNs)~\cite{long_shelhamer_fcn} has led to remarkable results in a variety of structured prediction tasks such as edge detection~\cite{gberta_2015_ICCV,DBLP:journals/corr/XieT15,DBLP:journals/corr/Kokkinos15} and semantic image segmentation~\cite{DBLP:journals/corr/ChenPKMY14,crfasrnn_iccv2015,DBLP:journals/corr/LiuLLLT15,DBLP:journals/corr/LinSRH15,noh2015learning,hong2015decoupled,DBLP:journals/corr/ChenYWXY15,Chen2016}.   Following this line of work, a recent method~\cite{ma2016going}, used FCNs for joint object segmentation and activity recognition in first person images using a two stream appearance and optical flow network with a multi-loss objective function. 



We point out that these prior methods~\cite{ma2016going,DBLP:conf/iccv/LiFR13,conf/cvpr/FathiRR11,conf/cvpr/RenG10,Boykov:2001:FAE:505471.505473} focus mainly on the RGB or motion cues, which is a very limiting assumption for an action-object task.  When interacting with an object, people typically position themselves at a certain distance and orientation relative to that object. Thus, 3D information plays an important role in action-object detection task. Unlike prior work, we integrate such 3D cues into our model for a more effective action-object detection. 

Additionally, the way a person positions himself during an interaction with an object, affects  where the object will be mapped in a first-person image. Prior methods~\cite{DBLP:journals/ijcv/LeeG15,DBLP:journals/corr/BertasiusPS15} assume that this will most likely be a center location in the image, which is a very general assumption. Instead, in this work, we introduce the first-person coordinate embedding to learn an action-object specific spatial distribution.

In this work, we show that our proposed additions are simple and easy to integrate into existing FCN framework, and yet they lead to a significant improvement in the action-object detection accuracy in comparison to all the prior methods.

\section{First-Person Action-Object RGBD Dataset}
\label{dataset_sec}

We use two stereo GoPro Hero 3 cameras with 100mm baseline to capture first-person RGBD videos as shown in Figure~\ref{arch}. The stereo cameras are synchronized manually and each camera is set to 1280$\times$960 with 100 fps. The fisheye lens distortion is pre-calibrated and depth image is computed by estimating disparities between cameras via dense image matching with dynamic programming. 

Two subjects participated in capturing their daily interactions with objects in activities such as cooking, shopping, working at their office, dining, buying groceries, dish-washing, and staying in a hotel room. 7 scenes were recorded and $4229$ frames with per-pixel action-objects were annotated by the subjects with GrabCut~\cite{Rother:2004:GIF:1015706.1015720}. 

In Table~\ref{table:dataset}, we provide a brief summary of our First Person Action-Object Dataset. The dataset consists of $7$ sequences, which capture various people's interactions with objects. In comparison to the existing first-person datasets such as GTEA Gaze+~\cite{Li_2015_CVPR}, which records a person's interactions only during a cooking activity, our dataset captures more diverse person's interaction with objects. This allows us to leverage common person-object spatial configurations and build a more general action-object model without constraining ourselves to a specific task, as is done in~\cite{Li_2015_CVPR}. In the experimental section, we will show that our model generalizes well on a variety of first-person datasets even if they contain previously unseen scenes, objects or activities.







\section{EgoNet}

In this section, we describe EgoNet, a predictive network model that detects action-objects from a first-person RGBD image. EgoNet is a two-stream FCN that holistically integrates visual appearance, head direction, and 3D spatial cues, and that is specifically adapted for first-person data via a first-person coordinate embedding. EgoNet consists of 1) an RGB pathway that learns object visual appearance cues; 2) a DHG pathway that learns to detect action-objects based on 3D depth and height measurements around the person; and a 3) a joint pathway that combines the information from both pathways, and which also incorporates our proposed first-person coordinate embedding to model a spatial distribution of action-objects in the first-person view. The detailed illustration of EgoNet's architecture is presented in Figure~\ref{arch}. We now explain each of EgoNet's components in more detail. 

%

\subsection{RGB Pathway} 

An action-object that stimulates person's visual attention typically exhibits a particular visual appearance. For instance, we are more likely to look at the objects that are colored brightly and stand out from the background visually. Thus, our model should detect visual appearance cues that are indicative of action-objects. To achieve this goal we use  DeepLab~\cite{DBLP:journals/corr/ChenPKMY14}, which is a fully convolutional adaptation of a VGG network~\cite{Simonyan14c}. DeepLab has been shown to yield excellent results on problems such as semantic segmentation~\cite{DBLP:journals/corr/ChenPKMY14}. Just like  the segmentation task, action-object detection also requires producing a per-pixel probability map. Thus, inspired by the success of a DeepLab system on semantic segmentation task, we adopt a pretrained DeepLab's network architecture as our RGB pathway.

\subsection{DHG Pathway} 

An action-object also possesses the following 3D spatial properties. It exhibits characteristic distance to the person due to anthropometric constraints, e.g., arm length. For example, when a man picks up a tuna can, his distance from the can is approximately 0.5m. The action-object also has a specific orientation relative to a person because of its design. For instance, when the person carries a cup, he holds it via the handle, which determines the pose of the cup with respect to that person. These 3D spatial properties are essential for predicting the action-objects.

Considering this intuition, we use our collected first-person stereo data to encode spatial 3D cues in a DHG (depth, height, and grayscale) image input. We use depth and height~\cite{gupta14rcnndepth} to represent the 3D environment around the person, and to handle the pitch movements of the head, i.e., the height information tells us about the orientation of the person's head with respect to 3D environment. In addition, the gray scale image is used to capture basic visual appearance cues. Note that we do not use full RGB channels so that the DHG pathway would focus more on depth and height cues than the visual appearance cues. This diversifies the information learned by the two pathways, which allows EgoNet to learn complementary action-object cues. In Figure~\ref{filters}, we visualize the activation values from the \textit{fc7} layer of RGB and DHG pathway averaged across all channels. Note that the two pathways learn to detect complementary action-object cues. Whereas RGB pathway detects objects that are visually prominent, DHG pathway has high activation values around the objects with a certain distance and orientation relative to the person.


\captionsetup{labelformat=empty}
\captionsetup[figure]{skip=5pt}

\begin{figure}
\centering

\myfigurefoursmallcol{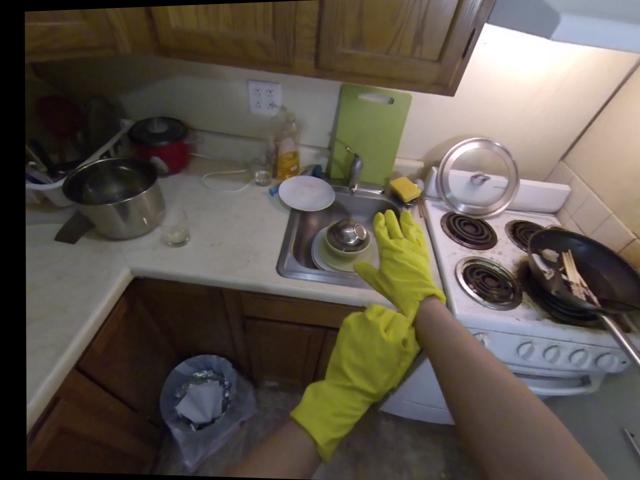}
\myfigurefoursmallcol{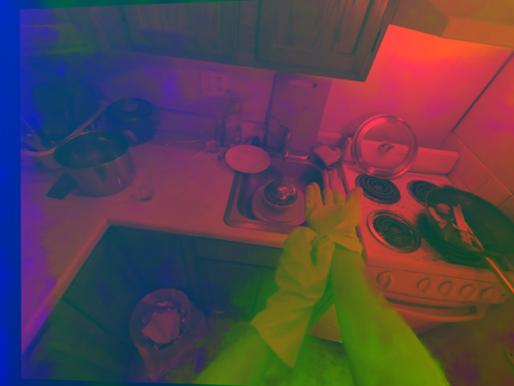}
\myfigurefoursmallcol{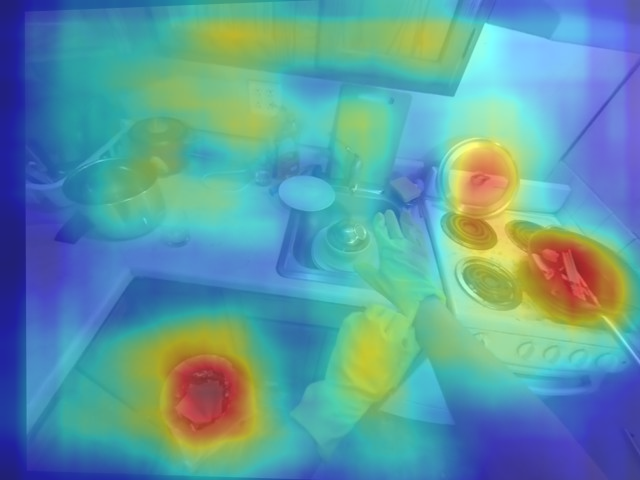}
\myfigurefoursmallcol{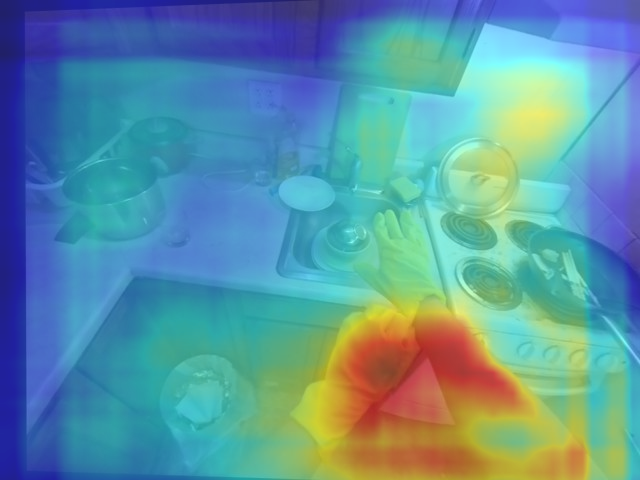}

\myfigurefoursmallcolcaption{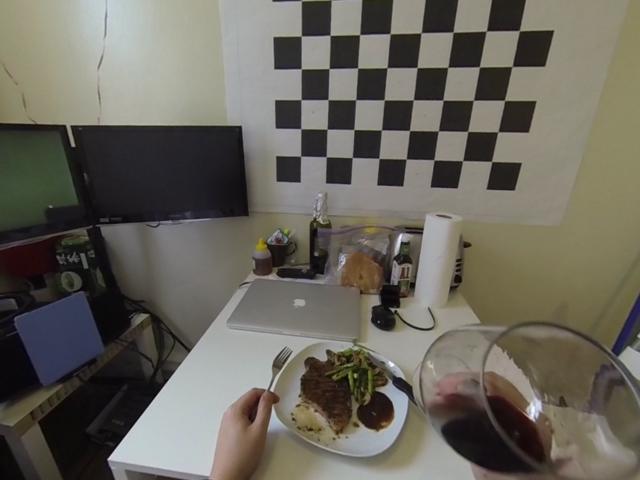}{RGB Input}
\myfigurefoursmallcolcaption{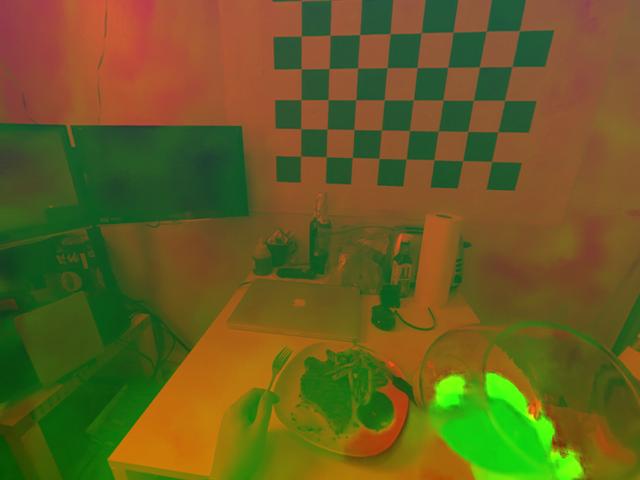}{DHG Input}
\myfigurefoursmallcolcaption{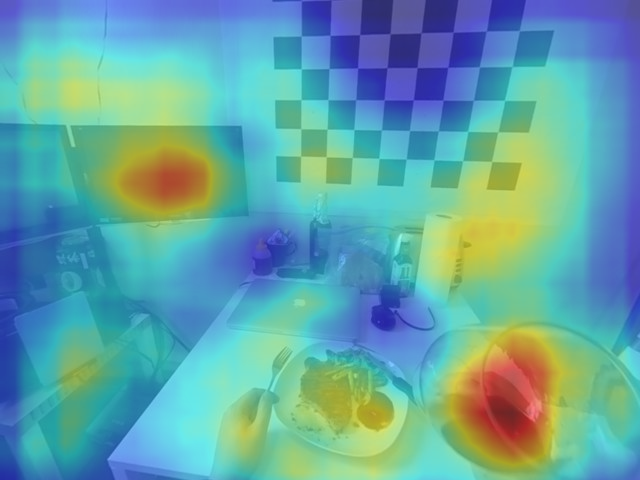}{RGB Pathway}
\myfigurefoursmallcolcaption{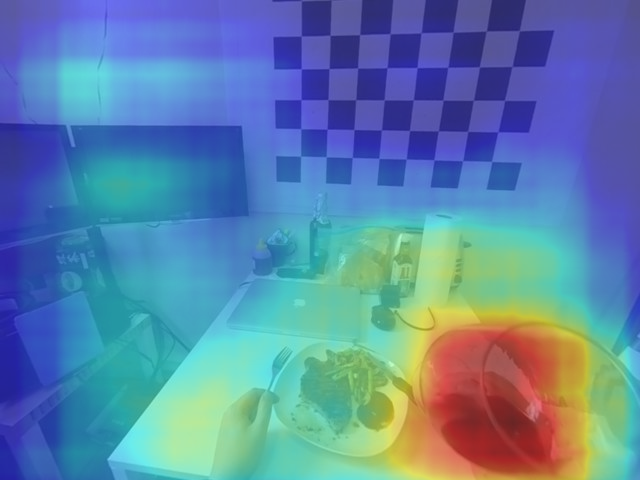}{DHG Pathway}


\captionsetup{labelformat=default}
\setcounter{figure}{2}
   \caption{The visualization of the \textit{fc7} activation values from the RGB and DHG pathways. The RGB pathway has higher activations around objects that stand out visually (e.g. a TV, a frying pan, a trash bin),  while the DHG pathway detects objects that are at a certain distance and orientation relative to the person (e.g. a wine glass, the gloves).\vspace{-0.5cm}}
    \label{filters}
\end{figure}

\captionsetup{labelformat=default}
\captionsetup[figure]{skip=10pt}


\captionsetup[figure]{labelformat=empty}
\captionsetup[figure]{skip=2pt}

\begin{figure*}[t]
\centering

\myfiguresixcol{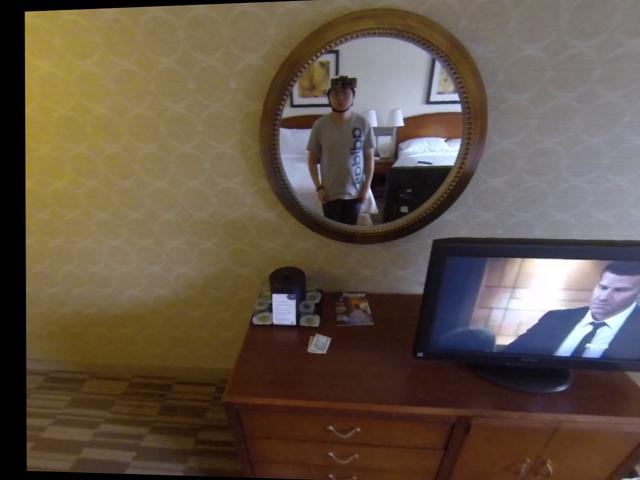}
\myfiguresixcol{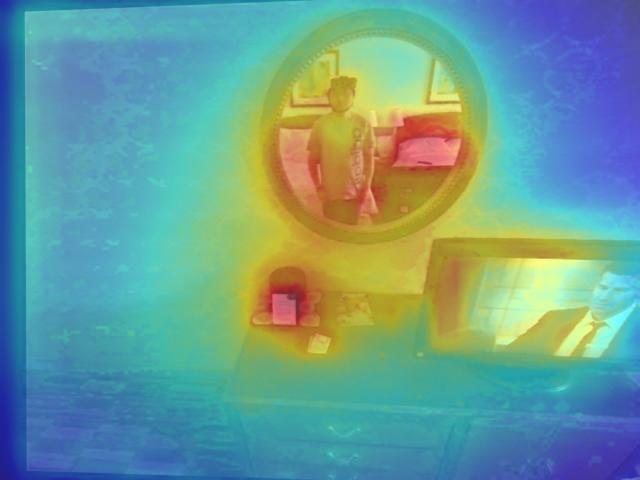}
\myfiguresixcol{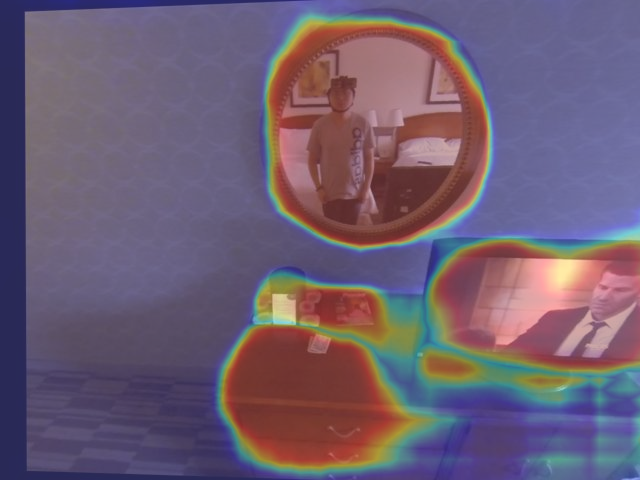}
\myfiguresixcol{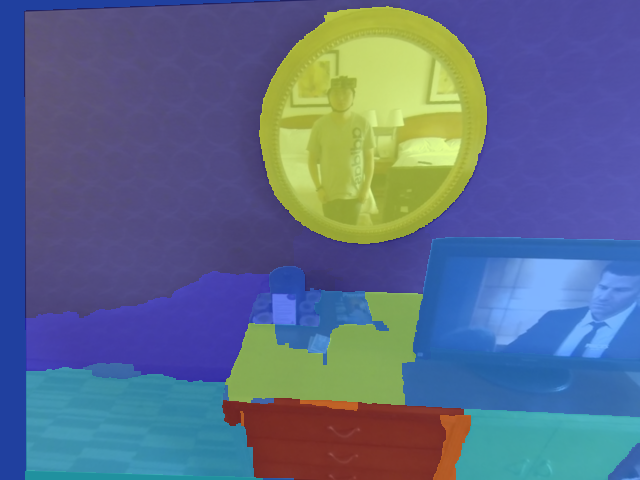}
\myfiguresixcol{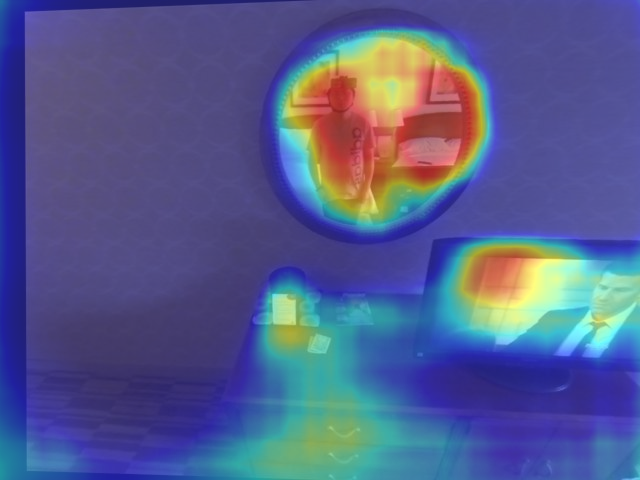}
\myfiguresixcol{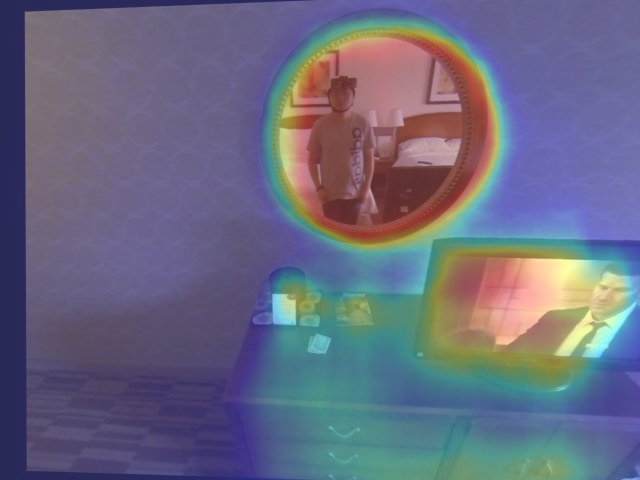}

\myfiguresixcolcaption{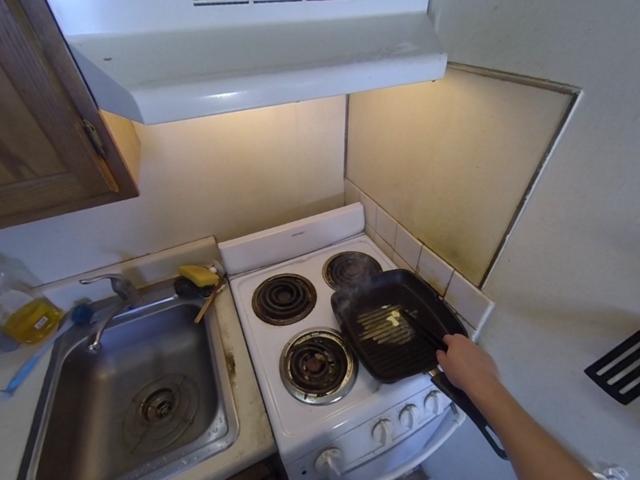}{RGB Input}
\myfiguresixcolcaption{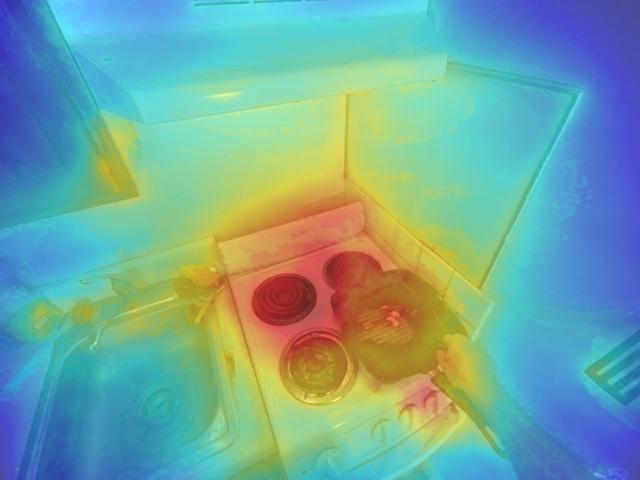}{Judd~\cite{Judd_2009}}
\myfiguresixcolcaption{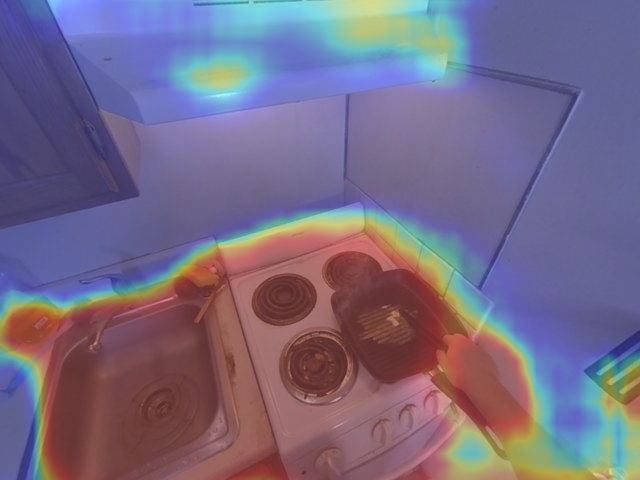}{DeepLab-obj~\cite{DBLP:journals/corr/ChenPKMY14}}
\myfiguresixcolcaption{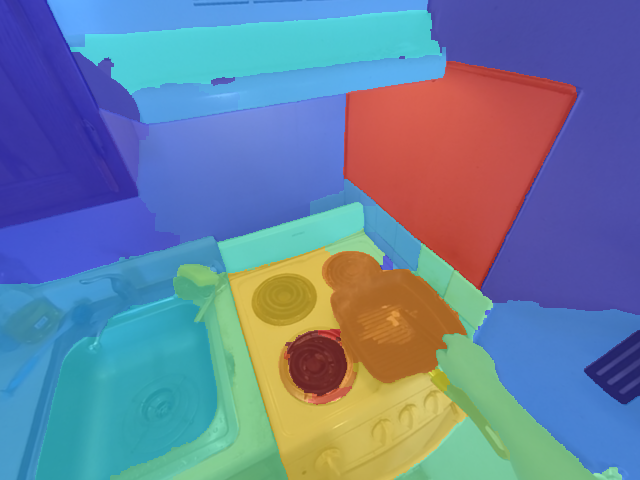}{salObj-depth~\cite{secrets2014li}}
\myfiguresixcolcaption{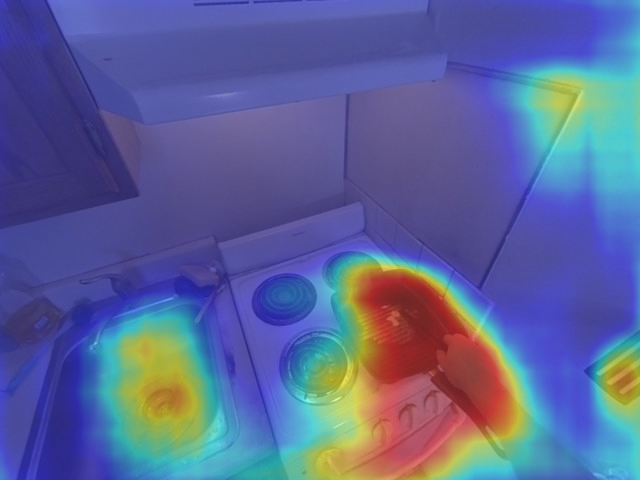}{DeepLab-dhg~\cite{DBLP:journals/corr/ChenPKMY14}}
\myfiguresixcolcaption{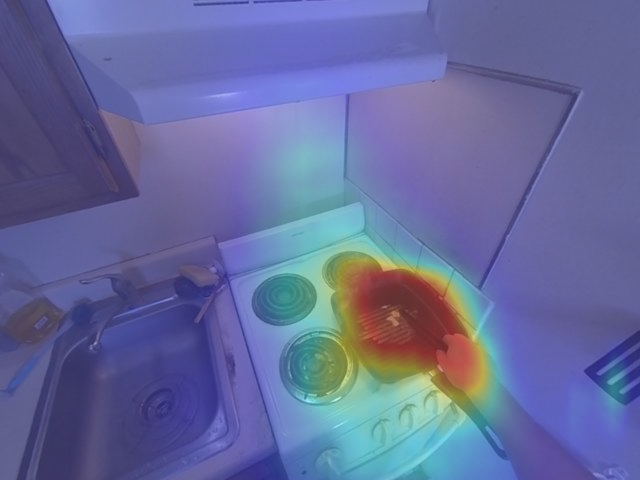}{EgoNet}
   
 \captionsetup{labelformat=default}  
 \setcounter{figure}{3}
\caption{An illustration of qualitative results on our dataset (the mirror and the fry pan are the action-objects). Unlike other methods, our EgoNet model correctly recognizes and localizes action-objects in both instances.\vspace{-0.4cm}}
    \label{egod_preds}
\end{figure*}


\subsection{Joint Pathway}

To combine the information from RGB and DHG pathways for action-object prediction we introduce a joint pathway. The joint pathway first concatenates $39\times39\times4096$ dimensional \textit{fc7} features from both RGB and DHG pathways to obtain a $39\times39\times 8192$ dimensional tensor. This concatenated \textit{fc7} tensor is then also concatenated with the downsampled $X \in \mathbb{R}^{39 \times 39}$ and $Y \in \mathbb{R}^{39 \times 39}$ first-person coordinates, which are obtained by generating $X,Y$ coordinate mesh-grids associated with every pixel in the original first-person image, and then downsampling these mesh-grids by a factor of $8$, which is how much the resolution of the output is reduced inside the FCN. Then, the input to the joint pathway is a $39\times39\times 8194$ dimensional tensor. 

Afterwards, we perform a first-person coordinate embedding, which consists of (1) using the first-person spatial coordinates in a standard convolution operation, and then (2) attaching another convolutional layer to blend the visual and spatial information in the new layer. The first-person coordinate embedding allows EgoNet to use visual and spatial features in conjunction, which we show is beneficial for an accurate action-object detection in first-person images. 


Intuitively the importance of first-person coordinate embedding can be explained as follows. The way a person positions himself during an interaction relative to an action-object, affects a location where such an action-object will be mapped in a first-person image. For instance, a laptop keyboard is often seen at the bottom of a first person image because we often look down at it while typing with our hands. Our proposed first-person coordinate embedding allows EgoNet to learn such an action-object spatial distribution, which is different than most prior work that assumes a universal object spatial prior (e.g. a center prior)~\cite{Li_2015_CVPR,DBLP:journals/ijcv/LeeG15}.

One may think that our proposed first-person coordinate embedding should have a minimal effect to the network's performance because traditional FCNs also incorporate certain amount of spatial information in its prediction mechanism. However, unlike traditional FCNs, EgoNet uses first-person coordinates as features directly in the 2D convolution operation, which forces EgoNet to produce a different convolutional output than traditional FCNs would. Despite the simplicity of our first-person coordinate embedding scheme, in our experiments, we show that it significantly boosts the action-object detection accuracy.

 \setlength{\tabcolsep}{1.5pt}

   \begin{table*}[t]
   \scriptsize
    \begin{center}
    \begin{tabular}{ c | c  c | c  c | c  c | c  c | c  c | c  c | c  c X c  c |}
    \cline{2-17}
    & \multicolumn{2}{ c |}{cooking}  & \multicolumn{2}{ c |}{dining} &  \multicolumn{2}{ c |}{grocery} & \multicolumn{2}{ c |}{hotel} & \multicolumn{2}{ c |}{desk work}  &  \multicolumn{2}{ c |}{shopping} & \multicolumn{2}{ c X}{dishwashing}  & \multicolumn{2}{ c |}{mean}\\
    \cline{2-17}
    &  \multicolumn{1}{ c }{MF} &  \multicolumn{1}{ c |}{AP}  &  \multicolumn{1}{ c }{MF}  &  \multicolumn{1}{ c |}{AP}   & \multicolumn{1}{ c }{MF} & \multicolumn{1}{ c |}{AP} &  \multicolumn{1}{ c }{MF} &  \multicolumn{1}{ c |}{AP}  &  \multicolumn{1}{ c }{MF}  &  \multicolumn{1}{ c |}{AP}  & \multicolumn{1}{ c }{MF} & \multicolumn{1}{ c |}{AP} &  \multicolumn{1}{ c }{MF} &  \multicolumn{1}{ c X}{AP}  &  \multicolumn{1}{ c }{MF}  &  \multicolumn{1}{ c |}{AP}\\ \cline{1-17}
        \multicolumn{1}{| c |}{DeepLab-Obj~\cite{DBLP:journals/corr/ChenPKMY14}}  & 0.091 & 0.033 & 0.181 & 0.084 & 0.052 & 0.020 & 0.158 & 0.054 & 0.225 & 0.105 & 0.110 & 0.044 & 0.077 & 0.021 & 0.128 & 0.051 \\ \hline
       \multicolumn{1}{| c |}{Judd~\cite{Judd_2009}} & 0.188 & 0.091 & 0.160 & 0.092 & 0.048 & 0.021 & 0.286 & 0.189 & 0.523 & 0.428 & 0.102 & 0.051 & 0.063 & 0.030 & 0.182 & 0.107\\ \hline
       \multicolumn{1}{| c |}{GBVS~\cite{Harel07graph-basedvisual}} & 0.213 & 0.113 & 0.188 & 0.097 & 0.043 & 0.014 & 0.216 & 0.128 & 0.499 & 0.487 & 0.098 & 0.047 & 0.124 & 0.063 & 0.197 & 0.136 \\ \hline
        \multicolumn{1}{| c |}{Handled+Viewed-Obj}  & 0.243 & 0.119 & 0.436 & 0.357 & 0.115 & 0.034 & 0.122 & 0.018 & 0.197 & 0.062 & 0.269 & 0.175 & 0.140 & 0.069 & 0.217 & 0.119 \\ \hline
         \multicolumn{1}{| c |}{DeepLab-RGB~\cite{DBLP:journals/corr/ChenPKMY14}} & 0.342 & 0.220 & 0.220 & 0.143 & 0.134 & 0.063 & 0.146 & 0.065 & 0.292 & 0.213 & 0.158 & 0.073 & 0.262 & 0.128 & 0.222 & 0.129 \\ \hline
         \multicolumn{1}{| c |}{AOP}  & 0.366 & 0.264 & 0.195 & 0.084 & 0.180 & 0.086 & \bf 0.394 & 0.222 & 0.421 & 0.327 & 0.137 & 0.074 & 0.267 & 0.178 & 0.280 & 0.176 \\ \hline
      \multicolumn{1}{| c |}{FP-MCG~\cite{APBMM2014}}  & 0.224 & 0.113 & 0.400 & 0.283 & \bf 0.243 & \bf 0.126 & 0.274 & 0.136 &  0.597 & 0.389 &  0.200 & 0.093 & 0.281 & 0.170 & 0.317 & 0.187 \\ \hline
       \multicolumn{1}{| c |}{DeepLab-DHG~\cite{DBLP:journals/corr/ChenPKMY14}} & 0.330 & 0.230 & 0.525 & 0.246 & 0.208 & 0.117 & 0.267 & 0.159 & 0.340 & 0.264 & 0.301 & 0.102 & 0.290 & 0.154 & 0.323 & 0.181 \\ \hline
       \multicolumn{1}{| c |}{salObj+depth~\cite{secrets2014li}} & 0.306 & 0.182 & \bf 0.551 & 0.451 & 0.188 & 0.105 & 0.361 & \bf 0.238 & 0.501 & 0.378 & \bf 0.404 & \bf 0.284 & 0.257 & 0.184 & 0.367 & 0.260 \\ \hline
        \multicolumn{1}{| c |}{\bf EgoNet}  & \bf 0.482 & \bf 0.415 & 0.509 & \bf 0.473 & 0.193 & 0.121 & 0.298 & 0.183 & \bf 0.643 & \bf 0.597 & 0.242 & 0.134 & \bf 0.406 & \bf 0.272 & \bf 0.396 & \bf 0.313 \\ \Xhline{3\arrayrulewidth}  
        \multicolumn{1}{| c |}{Subject 1} & 0.419 & - & 0.576 & - & 0.408 & - & 0.477 & - & 0.556 & - & 0.357 & - & 0.415 & - & 0.458 & - \\ \hline
        \multicolumn{1}{| c |}{Subject 2} & 0.453 & - & 0.551 & - & 0.327 & - & 0.444 & - & 0.516 & - & 0.518 & - & 0.436 & - & 0.464 & - \\ \hline
        \multicolumn{1}{| c |}{Subject 3} & 0.453 & - & 0.604 & - & 0.197 & - & 0.453 & - & 0.581 & - & 0.340 & - & 0.439 & - & 0.438 & - \\ \hline
        \multicolumn{1}{| c |}{Subject 4} & 0.506 & - & 0.631 & - & 0.358 & - & 0.489 & - & 0.579 & - & 0.371 & - & 0.407 & - & 0.477 & -\\ \hline
        \multicolumn{1}{| c |}{Subject 5} & 0.488 & - & 0.660 & - & 0.227 & - & 0.444 & - & 0.598 & - & 0.373 & - & 0.435 & - & 0.461 & - \\ \hline
    \end{tabular}
    \end{center}
    \vspace{-0.2cm}
  \caption{The quantitative results for action-object detection task on our first-person action-object RGBD dataset according to max F-score (MF) and average precision (AP) metrics. All methods except GBVS were trained on our dataset using a leave-one-out cross validation. The result indicates that EgoNet has the strongest predictive power with at least $2.9\%$ (MF) and $5.3\%$ (AP) gain over other methods. We also include our human study results, which suggest that human subjects achieve better action-object detection accuracy than the machines across most activities from our dataset.\vspace{-0.4cm}}
    \label{egod_table}
   \end{table*}

   \setlength{\tabcolsep}{1.5pt}

\subsection{Implementation Details}

We implement EgoNet using Caffe~\cite{jia2014caffe}. The RGB and DHG pathways are built upon DeepLab architecture~\cite{DBLP:journals/corr/ChenPKMY14} The entire EgoNet network is trained jointly for $3000$ iterations, at the learning rate of $10^{-6}$, momentum of $0.9$, weight decay of $0.0005$, batch size of $15$, and the dropout rate of $0.5$. To optimize the network, we used a per-pixel softmax loss with respect to the action-object ground truth. 

\section{Experimental Results}
\label{results}

In this section, we present quantitative and qualitative results of our EgoNet method on (1) our collected First Person Action-Object RGBD, (2) GTEA Gaze+~\cite{Li_2015_CVPR}, (3) Social Children Interaction~\cite{park_cvpr:2015}. Additionally, to gain a deeper insight into an action-object detection problem we also include an action-object human study where $5$ human subjects perform this task on our dataset.

To evaluate an action-object detection accuracy we use maximum F-score (MF), and average precision (AP) evaluation metrics, which are obtained by thresholding probabilistic action-object maps at small intervals and computing a precision and recall curve. Our evaluations provide evidence for four main conclusions:


\begin{itemize}
\item Our human-study indicates that humans achieve better action-object detection accuracy than the machines.
\item We also demonstrate that our EgoNet model outperforms all other approaches by a considerable margin on our First-Person Action-Object RGBD dataset.
\item Furthermore, we empirically justify the design choices for our EgoNet model.
\item Finally, we show that EgoNet performs well on the other novel first-person datasets.
\end{itemize}


\subsection{Action-Object Human Study}
\label{human_study_sec}

To gain a deeper insight into an action-object detection task, we conduct a human study experiment to see how well humans can detect action-objects from first-person images. We randomly select $100$ different first-person images from each of $7$ activities from our First-Person Action-Object RGBD dataset, and ask human subjects to identify a location of each action-object in such a first-person image. 

We use $100$ different images from each activity to keep the experiment's duration under an hour. Also, instead of collecting per-pixel or bounding box labels, we ask the subjects to identify action-objects by clicking at the center of an action-object, which is very efficient. We collect experimental action-object detection data from $5$ subjects.

%
%
%
%
%
%

  \begin{figure}[t]
 \centering
  \includegraphics[width=0.95\linewidth]{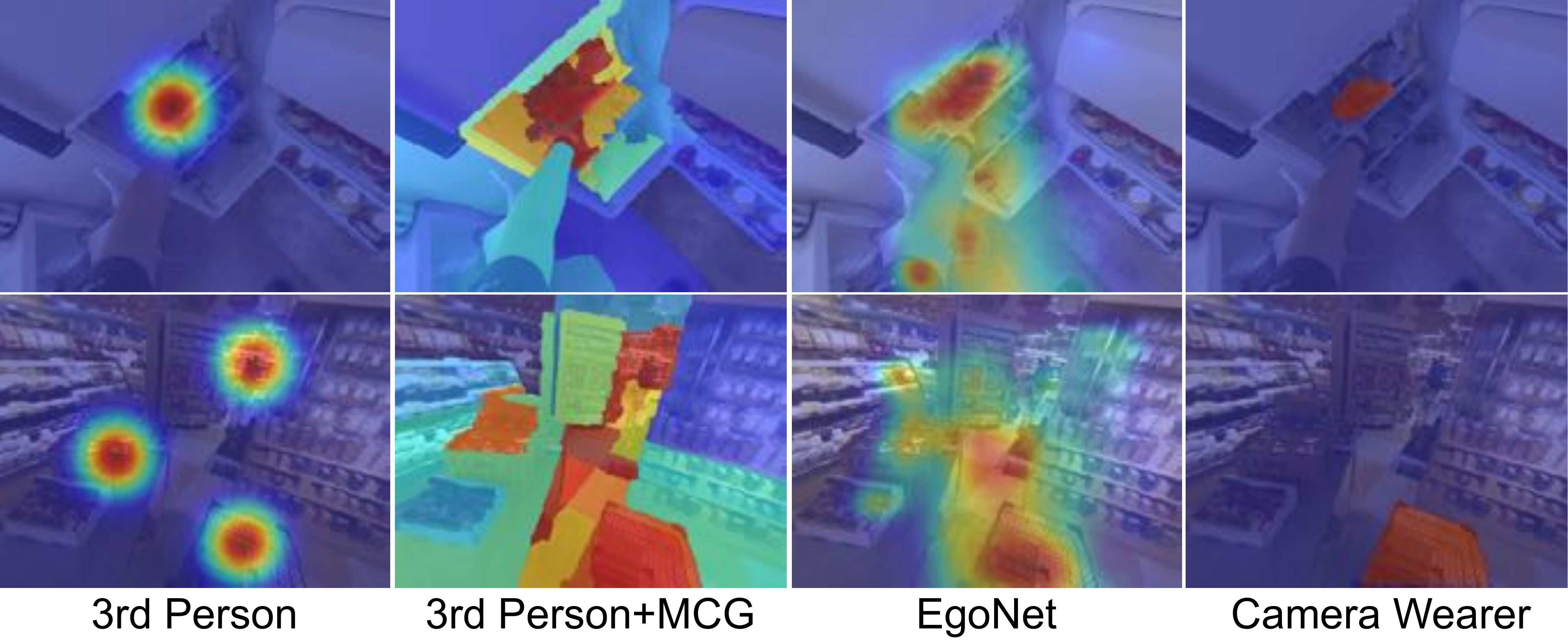}
 \captionsetup{labelformat=default}  
 \setcounter{figure}{4}
\caption{Qualitative human study results averaged across $5$ human subjects. In many cases, third-person human subjects detect action-objects correctly and consistently. However, some activities such as shopping makes this task difficult even for a human observer since he does not know what the camera wearer was thinking.\vspace{-0.4cm}}
    \label{human_study_fig}
 \end{figure}


To obtain full action-object segmentations from the points selected by the subjects, we place a Gaussian with a width of $60$ around the location of human selected point and project it on MCG~\cite{APBMM2014} regions as is done in~\cite{pinheiro:2015a}. We acknowledge that due to the use of Gaussian and the errors in the MCG algorithm, our scheme of obtaining per-pixel segmentations out of a single point may slightly degrade human subject results . However, even under current conditions a single experiment took about an hour or even more to complete. Thus, we believe that our chosen experiment conditions provided the best trade-off between the experiment duration and the quality of the action-object detection results provided by the subjects.

\setlength{\tabcolsep}{1.5pt}

   \begin{table*}[t]
   \scriptsize
    \begin{center}
    \begin{tabular}{ c | c | c | c | c | c | c | c | c | c | c | c | c | c | c X c | c |}
    \cline{2-17}
    & \multicolumn{2}{ c |}{breakfast}  & \multicolumn{2}{ c |}{pizza} &  \multicolumn{2}{ c |}{snack} & \multicolumn{2}{ c |}{salad} & \multicolumn{2}{ c |}{pasta}  &  \multicolumn{2}{ c |}{sandwich} & \multicolumn{2}{ c X}{burger}  & \multicolumn{2}{ c |}{mean}\\
    \cline{2-17}
    &  \multicolumn{1}{ c |}{MF} &  \multicolumn{1}{ c |}{AP}  &  \multicolumn{1}{ c |}{MF}  &  \multicolumn{1}{ c |}{AP}   & \multicolumn{1}{ c |}{MF} & \multicolumn{1}{ c |}{AP} &  \multicolumn{1}{ c |}{MF} &  \multicolumn{1}{ c |}{AP}  &  \multicolumn{1}{ c |}{MF}  &  \multicolumn{1}{ c |}{AP}  & \multicolumn{1}{ c |}{MF} & \multicolumn{1}{ c |}{AP} &  \multicolumn{1}{ c |}{MF} &  \multicolumn{1}{ c X}{AP}  &  \multicolumn{1}{ c |}{MF}  &  \multicolumn{1}{ c |}{AP}\\ \cline{1-17}
      \multicolumn{1}{| c |}{salObj+depth~\cite{secrets2014li}} & 0.192 & 0.113 & 0.233 & 0.147 & 0.303 & 0.225 & 0.191 & 0.112 & 0.202 & 0.117 & 0.346 & 0.248 & 0.211 & 0.108 & 0.240 & 0.153 \\ \hline
       	\multicolumn{1}{| c |}{DeepLab-RGB~\cite{DBLP:journals/corr/ChenPKMY14}} & 0.358 & 0.291 & 0.342 & 0.261 & 0.507 & 0.504 & 0.344 & 0.274 & 0.339 & 0.291 & 0.475 & 0.435 & 0.348 & 0.294 & 0.388 & 0.336 \\ \hline
        \multicolumn{1}{| c |}{Judd~\cite{Judd_2009}}  & 0.310 & 0.223 & 0.318 & 0.214 & 0.522 & 0.482 & 0.404 & 0.329 & 0.393 &  0.315 & 0.452 & 0.374 & 0.403 & 0.304 & 0.400 & 0.320 \\  \hline
        \multicolumn{1}{| c |}{FP-MCG~\cite{APBMM2014}}  & 0.364 & 0.281 & 0.456 & 0.380 & 0.486 & 0.399 & 0.401 & 0.310 & 0.411 &  0.346 & 0.511 & 0.454 & 0.321 & 0.207 & 0.422 & 0.340 \\  \hline
         \multicolumn{1}{| c |}{AOP}  & 0.403 & 0.307 & \bf 0.499 & \bf 0.443 & 0.488 & 0.380 & 0.337 & 0.211 & 0.408 &  0.273 & 0.472 & 0.375 & 0.369 & 0.268 & 0.425 & 0.322\\  \hline
         \multicolumn{1}{| c |}{DeepLab-DHG~\cite{DBLP:journals/corr/ChenPKMY14}} & 0.400 & 0.344 & 0.440 & 0.429 & 0.577 & 0.564 & 0.429 & 0.374 & 0.467 & 0.429 & 0.456 & 0.419 & 0.368 & 0.304 & 0.448 & 0.409 \\ \hline
         \multicolumn{1}{| c |}{\bf EgoNet }  & \bf 0.433 & \bf 0.357 & 0.475 & 0.383 & \bf 0.607 & \bf 0.568 & \bf 0.512 & \bf 0.450 & \bf 0.532 & \bf 0.505 & \bf 0.576 & \bf 0.486 & \bf 0.454 & \bf 0.358 & \bf 0.513 & \bf 0.443 \\ \hline
    \end{tabular}
    \end{center}
    \vspace{-0.2cm}
     \caption{Quantitative results on GTEA Gaze+ dataset for an action-object detection task. To test each model's generalization ability on GTEA Gaze+ dataset, we train each method \textbf{only} on our First-Person Action-Object RGBD dataset. Based on the results, we observe that EgoNet exhibits the strongest generalization power.\vspace{-0.4cm}}
    \label{gt_table}
   \end{table*}


In the bottom of Table~\ref{egod_table}, we provide human subject results according to the MF evaluation metric. Unlike in our other experiments, the action-object masks obtained by the human subjects were skewed towards the extreme probability values of $1$ and $0$, which made the AP metric less informative. Thus, we only used MF score in this case. 


Based on these results, we observe that in most cases, each of the $5$ subjects perform better than the machines. We also observe that the action-object detection results achieved by the human subjects are quite consistent across most of different activities from our dataset. This indicates that humans can perform action-object detection from first-person images pretty effectively, despite not knowing exactly what the camera wearer was thinking (but possibly predicting it).

In Figure~\ref{human_study_fig}, we also present some qualitative human study results, where we average the predictions across all $5$ human subjects. These results indicate that in some instances (second row), detecting action-objects is pretty difficult even for the human subjects since they do not know what the camera wearer was thinking. 


\subsection{Results on Our Action-Object RGBD Dataset}
\label{rgbd_ao_exp}
   

In Table~\ref{egod_table} we present the results on our First-Person Action-Object dataset, which contains $4247$ annotated images. All the results are evaluated according to the MF and AP metrics.  We include the following baseline methods in our comparisons: (1-2) GBVS~\cite{Harel07graph-basedvisual} and Judd~\cite{Judd_2009}: two bottom-up visual saliency methods; (3) FP-MCG~\cite{APBMM2014}: a multiscale object segmentation proposal method that was trained on our first-person dataset; (4) Handled+Viewed Object: our trained method that detects objects around hands, if no hands are detected it predicts objects near the center of an image, which is where a person may typically look at; (5) Action-Object Prior (AOP): the average action-object location mask obtained from our dataset; (6) DeepLab-Obj~\cite{DBLP:journals/corr/ChenPKMY14}: a DeepLab network trained for traditional object-segmentation on our dataset with 41 object classes; (7-8) DeepLab-RGB~\cite{DBLP:journals/corr/ChenPKMY14} and DeepLab-DHG~\cite{DBLP:journals/corr/ChenPKMY14}: a DeepLab network trained for action-object detection using RGB and DHG images as its inputs respectively; (9) salObj+depth~\cite{secrets2014li}: a salient object detection system, which we adapt to also handle a depth input. 

Note that all methods except for GBVS are trained on our dataset. The training is done using a leave-one-out cross validation as is standard. Based on the results in Table~\ref{egod_table}, we observe that EgoNet outperforms all baseline methods by at least $2.9\%$ (MF) and $5.3\%$  (AP). In Figure~\ref{egod_preds}, we also show our qualitative action-object results. Unlike other methods, EgoNet correctly detects and localizes action-objects in all cases.

  \begin{figure}[t]
 \centering
  \includegraphics[width=0.95\linewidth]{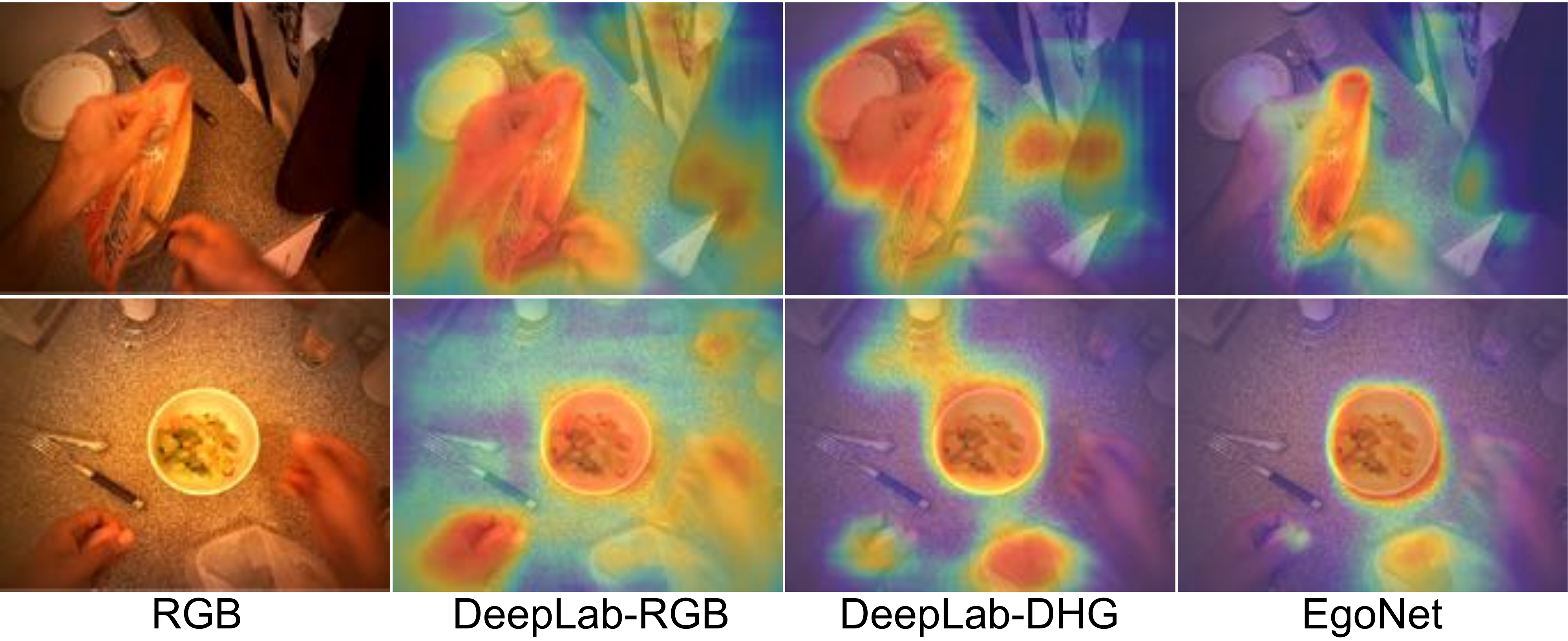}
 \captionsetup{labelformat=default}
 \setcounter{figure}{5}
     \caption{Qualitative results on GTEA Gaze+ dataset. EgoNet predicts action-objects more accurately and with better localization compared to DeepLab~\cite{DBLP:journals/corr/ChenPKMY14} based methods. \vspace{-0.6cm}}
     \label{gtea_fig}
 \end{figure}

\subsection{Analysis of EgoNet Architecture}
\label{analysis}

In this section, we quantitatively characterize the design factors of our EgoNet architecture on our dataset.



\textbf{Are Separate RGB and DHG Pathways Necessary?} An intuitive alternative to our EgoNet architecture is a single-stream network that concatenates RGB, DHG, and first-person coordinate inputs and feeds them through the network. We test such a baseline and report that it yields $0.249$ and $0.166$ MF and AP scores, which is significantly worse than $0.396$ (MF) and $0.313$ (AP) attained by our EgoNet model.






%

\textbf{What is the Contribution of RGB and DHG Pathways?} Our EgoNet model predicts action-objects based on the visual appearance and 3D spatial cues, which are learned in the separate RGB and DHG pathways. To examine how important each pathway is, we train two independent RGB and DHG single-stream networks (both with the first-person coordinate embedding). Whereas the RGB network obtains $0.363$ and $0.250$ MF and AP scores, the DHG network achieves $0.369$ and $0.208$ MF and AP results. These results indicate that the 3D spatial cues are equally or even more informative than the RGB cues.


\textbf{How Beneficial is the Joint Pathway?} We note that combining RGB and DHG pathways via the joint pathway yields $0.396$ and $0.313$ according to MF and AP metrics, which is a substantial improvement over the independent RGB and DHG networks, which achieve $0.363$ (MF), $0.250$ (AP), and $0.369$ (MF), $0.208$ (AP) scores respectively. This suggests, that RGB and DHG pathways learn complementary action-object information, and combining them via the joint pathway is beneficial. 

\textbf{Comparison with a Deeper Single Stream RGB Model.} We also include a single-stream network baseline that only uses RGB information, but that has about $17M$ more parameters than the RGB pathway in our EgoNet architecture. We report that such a baseline achieves $0.363$ (MF) score, which is identical to the RGB pathway in our EgoNet's design. Thus, these results indicate that simply adding more parameters to a single-stream network does not lead to better results.




\setlength{\tabcolsep}{1.5pt}
   \begin{table*}[t]
  \footnotesize
    \begin{center}
    \begin{tabular}{ c | c | c | c | c | c | c | c | c | c | c | c | c | c | c | c | c | c | c X c | c |}
    \cline{2-21}
    & \multicolumn{2}{ c |}{video 1}  & \multicolumn{2}{ c |}{video 2} &  \multicolumn{2}{ c |}{video 3} & \multicolumn{2}{ c |}{video 4} & \multicolumn{2}{ c |}{video 5}  &  \multicolumn{2}{ c |}{video 6} &  \multicolumn{2}{ c |}{video 7} & \multicolumn{2}{ c |}{video 8} & \multicolumn{2}{ c X}{video 9}  & \multicolumn{2}{ c |}{mean}\\
    \cline{2-21}
    &  \multicolumn{1}{ c |}{MF} &  \multicolumn{1}{ c |}{AP}  &  \multicolumn{1}{ c |}{MF}  &  \multicolumn{1}{ c |}{AP}   & \multicolumn{1}{ c |}{MF} & \multicolumn{1}{ c |}{AP} &  \multicolumn{1}{ c |}{MF} &  \multicolumn{1}{ c |}{AP}  &  \multicolumn{1}{ c |}{MF}  &  \multicolumn{1}{ c |}{AP}  & \multicolumn{1}{ c |}{MF} & \multicolumn{1}{ c |}{AP} &  \multicolumn{1}{ c |}{MF} & \multicolumn{1}{ c |}{AP} &  \multicolumn{1}{ c |}{MF} &  \multicolumn{1}{ c |}{AP}  &  \multicolumn{1}{ c |}{MF}  &  \multicolumn{1}{ c X}{AP} & \multicolumn{1}{ c |}{MF}  &  \multicolumn{1}{ c |}{AP}\\ \cline{1-21}
       \multicolumn{1}{| c |}{DeepLab-RGB~\cite{DBLP:journals/corr/ChenPKMY14}} & 0.113 & 0.062 & 0.125 & 0.069 & 0.116 & 0.073 & 0.132 & 0.073 & 0.089 & 0.044 & 0.093 & 0.046 & 0.113 & 0.060 & 0.090 & 0.036 & 0.160 & 0.082 & 0.115 & 0.061\\ \hline
       \multicolumn{1}{| c |}{DeepLab-DHG~\cite{DBLP:journals/corr/ChenPKMY14}} & 0.144 & 0.079 & 0.147 & 0.074 & 0.126 & 0.073 & 0.123 & 0.065 & 0.095 & 0.045 & 0.100 & 0.047 & 0.110 & 0.056 & 0.089 & 0.038 & 0.171 & 0.092 & 0.123 & 0.063\\ \hline
        \multicolumn{1}{| c |}{FP-MCG~\cite{APBMM2014}}  & 0.181 & 0.096 & 0.164 & 0.072 & 0.126 & 0.058 & 0.134 & 0.061 & 0.120 &  0.044 & 0.117 & 0.043 & 0.132 & 0.049 & 0.103 & 0.048 & 0.194 & 0.107 & 0.141 & 0.064 \\  \hline
         \multicolumn{1}{| c |}{AOP}  & 0.250 & 0.098 & 0.272 & 0.110 & 0.245 & 0.101 & 0.258 & 0.104 & \bf 0.257 &  0.101 & \bf 0.246 & 0.099 & \bf 0.240 & 0.096 & \bf 0.256 & 0.105 & 0.265 & 0.138 & 0.254 & 0.106\\  \hline     
         \multicolumn{1}{| c |}{\bf EgoNet }  & \bf 0.325 & \bf 0.236 & \bf 0.317 & \bf 0.217 & \bf 0.317 & \bf 0.211 & \bf 0.330 & \bf 0.225 & 0.246 & \bf 0.133 & 0.232 & \bf 0.142 & 0.237 & \bf 0.144 & 0.247 & \bf 0.140 & \bf 0.318 & \bf 0.212 & \bf 0.285 & \bf 0.185\\ \hline
    \end{tabular}
    \end{center}
    \vspace{-0.2cm}
    \caption{Quantitative results on Social Children Interaction dataset~\cite{park_cvpr:2015} for the visual attention prediction task. The dataset contains $9$ first-person videos of children performing activities such as playing cards games, etc. We note that to test each method's generalization ability, none of the methods were trained on this dataset. We show that EgoNet outperforms all the other methods on this dataset, which indicates strong EgoNet's generalization power.\vspace{-0.2cm}} 
    \label{children_table}
   \end{table*}

\textbf{Do First-Person Coordinates Help?} Earlier we claimed that using first-person coordinates in the joint pathway is essential for a good action-object detection performance. To test this claim, we train a network with an identical architecture as EgoNet except that it \textbf{does not} use first-person coordinates. Such a network yields $0.333$ and $0.232$ MF and AP scores, which is considerably lower than $0.396$ and $0.313$ MF and AP results produced by our proposed EgoNet model, that uses first-person coordinates. Such a big accuracy difference between the two models suggests that first-person coordinates play a crucial role in an action-object detection task.

%
%
%

  \begin{figure}[t]
 \centering
  \includegraphics[width=0.95\linewidth]{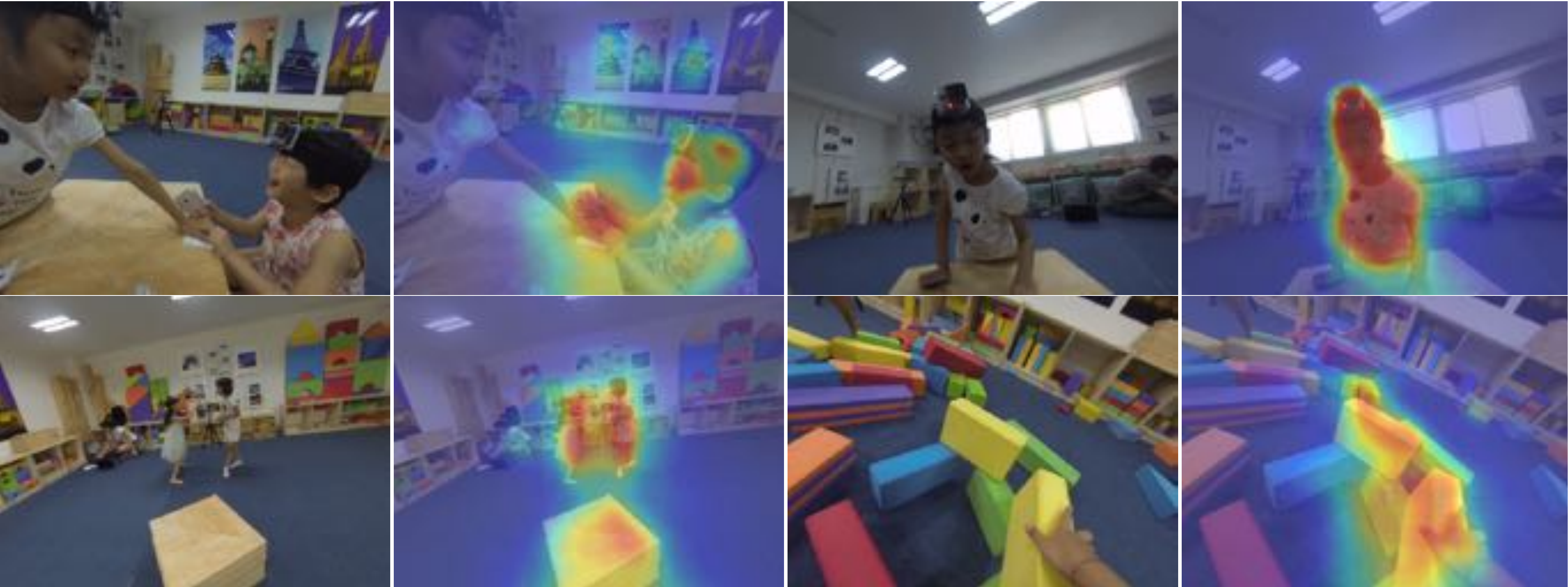}
 \captionsetup{labelformat=default}
 \setcounter{figure}{6}
     \caption{Our results on Social Children Interaction Dataset. Strong EgoNet's generalization power allows it to predict action-objects in a novel scenes, that contain previously unseen objects, and activities.\vspace{-0.5cm}}
     \label{children_preds}
 \end{figure}

\textbf{Is the Coordinate Embedding Useful?} In the previous section, we also claimed that the first-person coordinate embedding (see Fig.~\ref{arch}) is crucial for a good action-object detection accuracy. To test this claim we train a network with an identical architecture as EgoNet except that we remove the last layer before softmax loss (i.e. where the coordinate embedding is performed). We observe that the network that \textbf{does not} use the coordinate embedding produces a $0.313$ and $0.202$ MF and AP scores, which is considerably lower than $0.396$ and $0.313$ achieved by our full EgoNet model. 

\subsection{Results on GTEA Gaze+ Dataset} 
\label{gtea_exp}

To show strong EgoNet's generalization ability, in Table~\ref{gt_table}, we present our action-object detection results on the GTEA Gaze+ dataset~\cite{Li_2015_CVPR}, which consists of first-person videos that capture people cooking $7$ different meals. The dataset provides the annotations of objects that people are interacting with during a cooking activity. In comparison to our First-Person Action-Object RGBD dataset, GTEA Gaze+ contains many novel scenes, and many new object classes, which makes it a good dataset for testing EgoNet's generalization ability. Additionally, we note that, GTEA Gaze+ does not have depth information in the scene. Thus, we augment the dataset with depth predictions using~\cite{Depth2015Liu}. 


Note that to test each model's generalization ability, all the methods are trained \textbf{only} on our First-Person Action-Object RGBD dataset. Based on the results in Table~\ref{gt_table}, we can conclude that EgoNet shows the strongest generalization power: EgoNet achieves $0.513$ (MF) and $0.443$ (AP), whereas the second best method yields $0.448$ (MF) and $0.409$ (AP). We also include qualitative detection results from this dataset in Figure~\ref{gtea_fig}, which shows that our method detects action-objects more accurately and with much better localization than the DeepLab-FCN~\cite{DBLP:journals/corr/ChenPKMY14} baselines.




\subsection{Results on Social Children Interaction Dataset}
\label{child_exp}

Furthermore, In Table~\ref{children_table}, we present our results on Social Children Interaction dataset~\cite{park_cvpr:2015}, which includes $9$ first person videos of three children playing a card game, building block towers, and playing hide-and-seek. The dataset consists of $2189$ frames that are annotated with the location of children's attention location~\cite{park_cvpr:2015}. To evaluate all methods on this dataset, we place a fixed size Gaussian around the ground truth attention location, and use it as our ground truth mask to evaluate the results according to MF and AP metrics. Similar to GTEA Gaze+ dataset, this dataset only contains RGB images so we complement it with the depth predictions using the method in~\cite{Depth2015Liu}.

To test the generalization power, we train each method \textbf{only} on our First-Person Action-Object RGBD dataset, and then test it on  Social Children Interaction dataset. We report that EgoNet achieves $0.285$ and $0.185$ MF and AP, whereas the second best method yields $0.254$ (MF) and $0.106$ (AP) scores.  We also illustrate several qualitative predictions in Figure~\ref{children_preds}. 



 
\subsection{Visualizing the Detected Action-Objects in 3D}
 
In Figure~\ref{3d_ao}, we also present a 3D visualization of the detected action-objects from our RGBD dataset. The recovered 3D spatial layout of the detected action-objects could be used to teach robots how to grasp and manipulate action-objects based on how humans interacted with these objects.

 \begin{figure}[t]
\centering
  \includegraphics[width=0.75\linewidth]{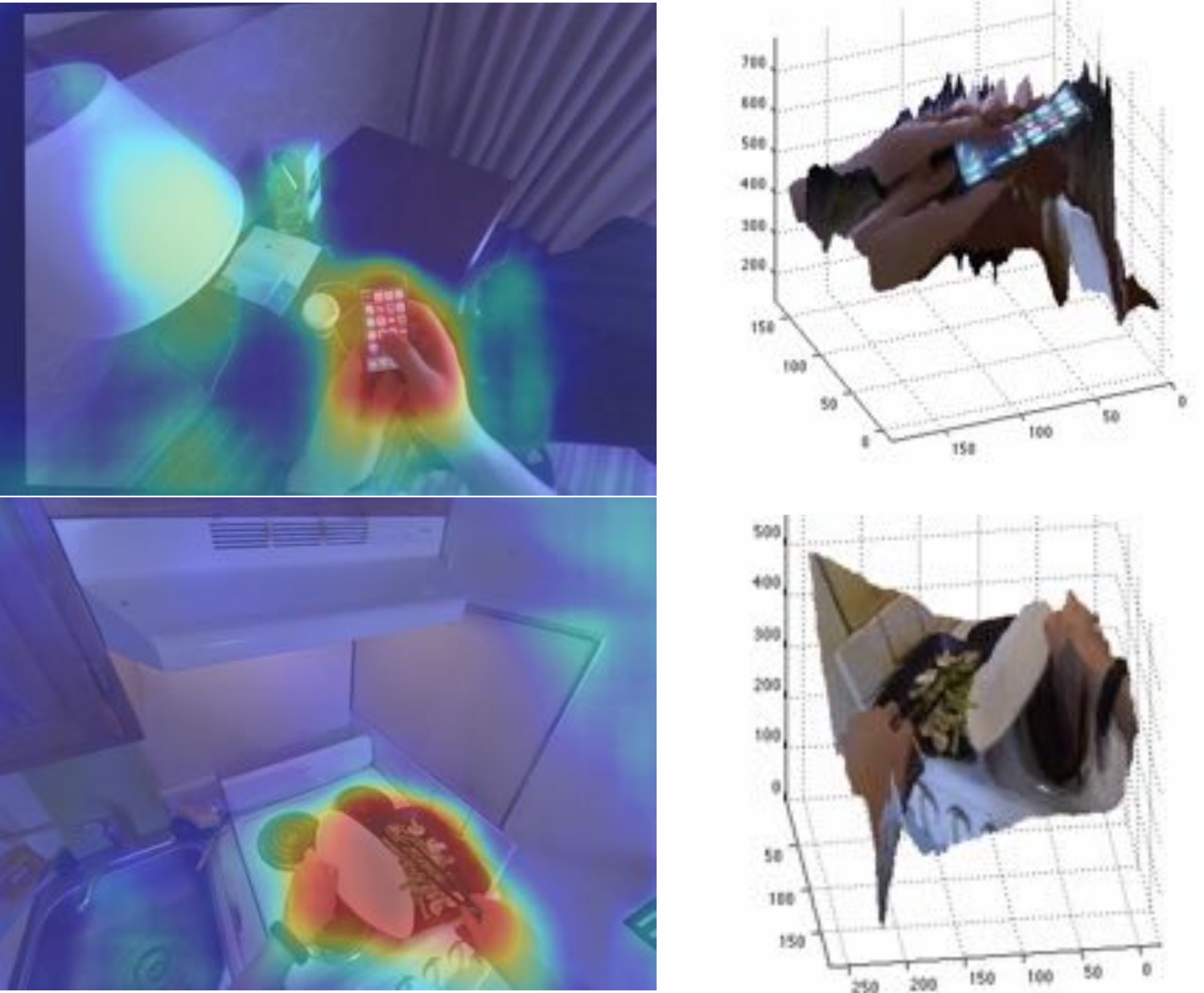}
   \captionsetup{labelformat=default}  
 \setcounter{figure}{7}
    \caption{A 3D visualization of the detected action-objects from our RGBD dataset (\textbf{Top}: using a phone, \textbf{Bottom}: cooking a meal). The details of how humans interacted with these objects could be used to teach robots how to manipulate such objects.\vspace{-0.5cm}}
    \label{3d_ao}
\end{figure}

\section{Conclusions}

In this work, we use a concept of an action-object to study a person's visual attention and his motor actions from a first-person visual signal. To do this, we introduce EgoNet, a two-stream network that holistically integrates visual appearance, head direction, 3D spatial cues, and that also employs first-person coordinate embedding for an accurate action-object detection from first-person RGBD data. Our EgoNet leverages common person-object spatial configurations, which allows it to predict action-objects without an explicit adaptation to a specific task as is done in prior work~\cite{Li_2015_CVPR,conf/cvpr/RenG10,ma2016going}. We believe that EgoNet's predictive power and its strong generalization ability makes it well suited for the applications, that require robots to understand how a person interacts with various objects, and using that information for assisting people.

\bibliographystyle{plain}
\footnotesize{
\bibliography{gb_bibliography_v2,bib_hs_v2}}

\end{document}